\documentclass[runningheads]{llncs}
\usepackage[T1]{fontenc}
\usepackage{graphicx}
\usepackage{amsmath}
\usepackage{amssymb}
\usepackage{bbm}
\usepackage{textcomp}
\usepackage{xcolor}
\usepackage{multirow}
\usepackage{enumitem}
\usepackage{bm}
\usepackage{booktabs}
\usepackage{algorithm, algorithmic}
\usepackage{subcaption}
\usepackage{cite}
\usepackage{url}
\usepackage[misc]{ifsym}

\def\done{\hspace*{\fill} \rule{1.8mm}{2.5mm} \\}

\begin{document}
\title{Contextual Bandit with Herding Effects: \\
Algorithms and Recommendation Applications}
\titlerunning{Contextual Bandit with Herding Effects}

\author{Luyue Xu\inst{1} \and
Liming Wang\inst{1} \and
Hong Xie\inst{2,3} \textsuperscript{(\Letter)} \and
Mingqiang Zhou\inst{1}}

\authorrunning{L. Xu Author et al.}

\institute{Chongqing University \and
University of Science and Technology of China 
\and State Key Laboratory of Cognitive Intelligence
\\
\email{xiehong2018@foxmail.com}
}

\maketitle

\begin{abstract}
Contextual bandits serve as a fundamental algorithmic framework for optimizing recommendation decisions online. Though extensive attention has been paid to tailoring contextual bandits for recommendation applications, the "herding effects" in user feedback have been ignored. These herding effects bias user feedback toward historical ratings, breaking down the assumption of unbiased feedback inherent in contextual bandits. This paper develops a novel variant of the contextual bandit that is tailored to address the feedback bias caused by the herding effects. A user feedback model is formulated to capture this feedback bias. We design the TS-Conf (Thompson Sampling under Conformity) algorithm, which employs posterior sampling to balance the exploration and exploitation tradeoff. 
We prove an upper bound for the regret of the algorithm, revealing the impact of herding effects on learning speed. 
Extensive experiments on datasets demonstrate that TS-Conf outperforms four benchmark algorithms. 
Analysis reveals that TS-Conf effectively mitigates the negative impact of herding effects, resulting in faster learning and improved recommendation accuracy.

\end{abstract}

\section{Introduction}

Contextual linear bandit  is an important sequential decision making framework 
for information retrieval applications \cite{Glowacka2019}.     
It is also applied to optimize news recommendations
\cite{li2010contextual,semenov2022diversity}, 
movie recommendations
\cite{pilani2021contextual,rao2020contextual}, 
advertising 
\cite{yang2016dynamic,srikanth2023dynamic}, etc.  
Recently, a number of variants of contextual linear bandits 
were proposed to capture important factors of information retrieval applications.  
Such as the conversational contextual bandit 
which captures the contextual linear bandit to capture 
conversational feedbacks in recommendation applications \cite{zhang2020conversational}, 
the impatient contextual bandits 
which captures feedback delay 
in recommendation applications \cite{Mcdonald2023},   
contextual budgeting bandit which captures the multi-agent nature 
the budget allocation in 
online advertising \cite{Han2021}, etc.   

This paper tackles a critical challenge in the field of recommendation systems: the herding effects in user feedback
\cite{adomavicius2016understanding,xie2020robust,wang2014quantifying}.
Randomized controlled experiments \cite{adomavicius2016understanding,muchnik2013social} 
proved the existence of herding effects.    
The herding effects states that users are conformed to historical ratings 
of a product when they are assigning ratings.  
We model the valuation or true preference 
of a user toward an item as a product of 
the item's feature vector and the user's preference vector.  
This valuation is unobservable, but a linear combination 
of the valuation and the historical rating is observable.    
The weights of this linear combination captures the strength of herding effects 
serves as a confounder and it is unobservable. 
This paper presents the first attempt to 
capture herding effects in contextual linear bandit, 
and we aim to reveal fundamental understandings on  
the impact of this unobservable confounder 
on balancing the exploration vs. exploitation tradeoff 
for the online recommendation task. 
Our contribution is the following. 

\begin{itemize}
	\item[\textbullet] 
        We propose a model to quantify herding effects, where a user's feedback is influenced by both its inherent reward and historical feedback. 
	\item[\textbullet] 
        We develop the TS-Conf algorithm, utilizing Thompson Sampling to balance exploration and exploitation effectively. We provide a regret upper bound for the algorithm, highlighting how herding effects, modulated by the conformity factor, affect learning efficiency.
	\item[\textbullet] 
        Extensive experiments on four public datasets demonstrate the 
        sublinear regret of the proposed TS-Conf algorithm, 
        and its superior performance over three baselines. 
\end{itemize}

\section{Related Work}
Contextual linear bandits serve as a fundamental sequential decision making framework 
for information retrieval applications advertising, recommendation, etc \cite{Glowacka2019}.    
A number of variants of contextual linear bandits 
were proposed to capture important factors of information retrieval applications \cite{wang2023efficient,xia2022multi,Xia2023,zuo2022hierarchical}.  
Conversational contextual bandit tailors the contextual linear bandit to capture 
conversational feedbacks in recommendation applications \cite{zhang2020conversational}.  
The contextual budgeting bandit extend the contextual linear bandit to the 
multi-agent for the purpose of studying the budget allocation in 
online advertising \cite{Han2021}. 
The key difference to the above work is that our model captures 
the well-known herding effects in feedback.  
The new technical challenge is that this herding effects 
leads to confounded feedback with spurious correlation.   

Through controlled experiments\cite{adomavicius2016understanding,muchnik2013social,salganik2006experimental}, 
some researchers identified a rating bias influenced by 
historical ratings, observing that users tend to 
give higher ratings after being exposed to 
higher historical ratings, a behavior termed as herding effects. 
Wang \textit{et al.} \cite{wang2014quantifying} introduced an additive generative-based model designed to quantify herding effects. While it can capture the pattern of herding effects, it lacks the neatness required for analytical studies of evolving dynamics of aggregate ratings under herding effects. 
Krishnan \textit{et al.} \cite{krishnan2014methodology} developed a polynomial regression-based model to quantify herding effects. 
Xie \textit{et al.} \cite{xie2020robust} proposed a neater linear model for herding effects 
and supported analytical studies on 
the evolving dynamics of aggregate ratings. 
Other notable psychological effects that lead to biased feedback include 
assimilate and contrast effects \cite{xie2023probabilistic},  
persuasion effects \cite{xie2021understanding}, etc.  
These works focus on rating prediction and are built on the matrix factorization framework.  
Unlike them, we consider the online decision setting built on the contextual bandit. 

A number of works studied bandit learning with biased feedback.  
Bareinboim \textit{et al.} \cite{bareinboim2015bandits} extended the reward model of 
multi-armed bandits such that an unobserved 
confounder influences the reward.  
Kallus \textit{et al.} \cite{kallus2018instrument} developed the reward model 
of multi-armed bandits such that an unobserved 
instrumental variable influences the reward. 
Different from them, our work is built on the contextual bandit. 
Maniu \textit{et al.} \cite{maniu2020bandits}  extended the linear bandits to 
capture social influence bias. 
In particular, they consider the setting that 
the preference vectors of users are influenced by their friends and evolve over time.  
Difference from their work, we apply the confounder model 
to capture herding effects.  
Tennenholtz \textit{et al.} \cite{tennenholtz2021bandits} studied linear contextual bandits 
with access to a large, confounded, offline dataset sampled from some fixed policy. 
Unlike their work, we consider the setting 
where an unobserved confounder influences 
the online reward. 
Sen \textit{et al.} \cite{sen2017contextual} studied 
stochastic contextual bandits with a latent low-dimensional confounder.  
The confounder is discrete and models the mood of users. 
Unlike them, our work considers a continuous confounder 
and the confounder models the strength of herding effects. 

\section{Model}
\subsection{The Sequential Decision Framework}
We consider the 
sequential decision problem 
as one where the 
decision-maker makes decisions 
over a finite number 
of $T\in\mathbb{N_+}$ rounds. 
The set of actions used 
in the decision-making process 
is fixed to be a finite 
set $\mathcal{A}\subset\mathbb{N_+}$, 
where $|\mathcal{A}|<\infty$. 
Consider a scenario 
where the decision-maker acts 
as a movie recommendation system, 
and $\mathcal{A}$ is the 
set of movies under consideration. 
In each round 
$t \in[T]\triangleq \{{1, . . . , T}\}$, 
the decision-maker 
is presented with a 
finite set of choices 
$\mathcal{A}_t\subseteq \mathcal{A}$ 
and $|\mathcal{A}_t|=K$, 
from which it must 
choose one action 
${A_t}\in[K] \triangleq \{ {1, . . . , K}\}$ 
to the user. 
The user then receives  
the expected preference reward 
$\mathbb{E}[R_t(A_t)]$ 
(positive or negative preference) 
for the recommending action, 
which is unobservable 
to the decision-maker. 
Based on the reward $\mathbb{E}[R_t(A_t)]$, 
each user provides  
feedback $V_t(A_t)\in\mathcal{R}$ 
about the action $A_t$ to decision-maker, 
where $\mathcal{R}\subset\mathbb{R}$. 
The application defines the metric for quantifying $V_{t}(A_t)$.
In movie recommendation applications, 
$V_{t}(A_t)$ models the 
user's rating of the movie. 
\subsection{The User Feedback Model}
\noindent \textbf{Contextual reward model.}
In our study, we focus on the use of contextual features to evaluate the rewards users receive from recommended items. 
For each action $a\in\mathcal{A}$, 
there is a feature vector $\boldsymbol{x}_a\in\mathbb{R}^d$ 
associated with it 
that captures contextual information 
between the user 
and the action 
with $d\in \mathbb{N}_+$. 
The preference vector 
of the user, linked to $\boldsymbol{x}_a$, 
is represented 
as $\boldsymbol{\theta}\in\mathbb{R}^d$. 
It should be emphasized that 
the decision maker 
has information about 
the observed context $\boldsymbol{x}_a$, 
$\forall a \in \mathcal{A}$, 
while $\boldsymbol{\theta}$ remains unknown. 
Given $\boldsymbol{x}_a,\boldsymbol{\theta}$,
in all round $t$, 
we consider the expected preference reward 
of the action $a$ from the user, modeled by: 
\begin{equation}
	\mathbb{E}[R_t(a)]=
	{\boldsymbol{\theta}^{\mathrm{T}}\boldsymbol{x}_a}
	\label{equq_reward}.
\end{equation} 
It is important to note that the decision maker cannot observe the reward at each recommendation round. However, after receiving the recommended action, users provide biased feedback. The goal of the decision maker is to maximize the expected cumulative rewards based on this biased user feedback.

\noindent \textbf{User feedback model.} 
We describe user’s feedback depends
on both expected reward formed by the user 
and the historical
feedback of actions. Specifically, 
in applications where ratings 
serve as feedback, 
the historical feedback for item $a$ is
determined by the rating content,  
which is known to the decision maker.
In each round $t$, 
let $h_{t,a}\in \mathbb{R_+}$ 
denote the historical feedback of action $a$. 
Let $\mathbb{E}[R_t(a)]$ represent  
the user's expected reward. 
According to the historical feedback $h_{t,a}$
and the expected reward 
model $\mathbb{E}[R_t(a)]$, 
we define that at time $t$, 
the user’s feedback to action $a$ is
modeled as:
\begin{equation}
\begin{aligned}
V_{t}(a)
&=
\alpha h_{t,a}+(1-\alpha)
\mathbb{E}[R_t(a)]+\eta_{t,a} =
\alpha h_{t,a}+(1-\alpha)
\boldsymbol{\theta}^{\mathrm{T}} 
\boldsymbol{x}_{a}+\eta_{t,a},
\end{aligned}
\label{eq:model:feedback}
\end{equation}
where $\alpha \in [0,1]$ denotes the 
conformity tendency of the user, 
which can be interpreted as the weights 
that balance between the historical rating $h_{t,a}$
and the expected reward $\mathbb{E}[R_t(a)]$. 
It should be noted that 
the decision maker has no 
idea about the exact value of $\alpha$.
$ {\eta}_{t,a} \in \mathcal{R}$ 
represents the stochastic noise 
caused by environmental. 
Let $f({\eta},\sigma_a)$ represent 
the probability density function 
of $ {\eta}_{t,a}$,
where $\sigma_a$ stands for 
the unknown standard deviation 
of $ {\eta}_{t,a}$ to the decision maker. 
The standard deviation $\sigma_a$ 
is also unknown to the decision maker. 
\subsection{Sequential Decision Making Model}
Based on the feedback model, we use $V_t(A_t)$ 
to study the unknown parameters of 
the user feedback and 
to get the user's preference estimate. 
To provide a clearer explanation, 
we consolidate all unknown 
parameters within $V_t(A_t)$ into 
$\bm{\Psi} \triangleq 
[\boldsymbol{\theta}, \alpha, \boldsymbol{\sigma}]$. 
Here, $\boldsymbol{\sigma} \triangleq 
[\sigma_1, . . . , \sigma_{|\mathcal{A}|}]$ represents 
the noise standard deviation 
for each action in the action set.
After round $t$, 
the decision-maker possesses 
information regarding the 
decision action $A_t$, 
the associated historical feedback $h_{t,A_t}$, the feedback $V_t(A_t)$, 
and the observed context $\boldsymbol{x}_{A_t}$.
Let $\mathcal{H}_t$ represent the decision-making
history up to decision round $t$ as  
$
\mathcal{H}_t
\triangleq
\{[A_1,h_{1,A_1},\boldsymbol{x}_{A_1},V_1(A_1)], \\
\ldots,[A_t,h_{t,A_t},
\boldsymbol{x}_{A_t},V_t(A_t)]\}
$.
In the $t$-th round 
of decision-making, 
the decision-maker must 
base decisions on the 
history $\mathcal{H}_{t-1}$ of 
the preceding $t-1$ rounds. 
Therefore, we propose using a 
sequential decision-making algorithm 
that leverages historical dependencies. 
Specifically, this algorithm maps the decision 
history to the current round’s 
action probability distribution 
$\mathcal{F}(\mathcal{H}_{t-1})$. 
The action $A_t$ is consequently 
generated from this distribution, 
expressed as $A_t\sim\mathcal{F}(\mathcal{H}_{t-1})$.
If the distribution 
$\mathcal{F}(\mathcal{H}_{t-1})$ targets a 
single action without variance, we have a 
deterministic scenario. To measure the effectiveness 
of a history-dependent algorithm with the 
probabilistic model $\mathcal{F}$, we present 
the regret function below: 
\begin{equation}
\begin{aligned}
R_{T}(\mathcal{F};\bm{\Psi})
\triangleq& 
\sum_{t=1}^{T} \max _{a \in \mathcal{A}_{t}} 
\mathbb{E}[{R}_{t}(a; \bm{\Psi})] - \mathbb{E}\left[
R_{t} \left(A_{t}\right) \mid 
\bm{\Psi}, A_{t} 
\sim \mathcal{F}\left(\mathcal{H}_{t-1}\right)
\right],
\end{aligned}
\label{equa4}
\end{equation}
where 
$\mathbb{E}[{R}_{t}(a; \bm{\Psi})]$ 
represents the expected 
reward for action $a$ 
under parameter $\bm{\Psi}$. 
A decision maker might have prior knowledge about the preference vector \(\boldsymbol{\theta}\), the conformity tendency \(\alpha\), and the standard deviation \(\boldsymbol{\sigma}\). We represent this prior knowledge using prior distributions over the parameters, expressed as \(p(\boldsymbol{\theta})\), \(p(\alpha)\), and \(p(\sigma_a)\) for each \(a \in \mathcal{A}\). We specifically consider scenarios where \(\boldsymbol{\theta}\), \(\alpha\), and \(\sigma_a\) independently arise from their respective prior distributions. This relationship is captured by the equation \(p(\bm{\Psi}) = p(\boldsymbol{\theta}) p(\alpha) \prod_{a\in \mathcal{A}} p(\sigma_a)\). The decision maker's goal is to design a sequential decision-making algorithm based on historical data that minimizes the regret.


\section{Algorithm}
\subsection{Algorithm Design}
In the context of accumulating decisions by round $t$, 
we articulate the model's parameters, 
still to be inferred, as 
$\bm{\Psi} = [\boldsymbol{\theta}, \alpha, \boldsymbol{\sigma}]$. 
The calculation for their posterior distribution, 
denoted $p\left(\bm{\Psi} \mid \mathcal{H}_{t}\right)$, 
is delineated in an ensuing lemma.

\noindent \textbf{Lemma 1.}
Suppose the probability density function of the noise has the parametric form 
$f(\cdot, \sigma)$, where $\sigma$ controls the tail property. Given the decision history 
$\mathcal{H}_{t}$ up to round $t$, 
the  posterior distribution 
$p\left(\bm{\Psi} \mid \mathcal{H}_{t}\right)$
can be derived as: 
\begin{equation}
\label{eque_posterior}
\begin{aligned}
p\left(\bm{\Psi} \mid \mathcal{H}_{t}\right) =& 
\frac{p(\bm{\Psi})}{C}
\times \left[ \prod\nolimits_{\tau=1}^{t-1} \prod\nolimits_{a\in\mathcal{A}_{\tau}} \left[f \left( \eta_{\tau,a}, \sigma_{a} \right)\right]^{\mathbbm{1}_{\left\{A_{\tau}=a\right\}}} \right],
\end{aligned}
\end{equation}
\begin{equation}
\label{eta_function}
\eta_{\tau,a}=
V_{\tau}(a)-
\alpha h_{\tau,a}-(1-\alpha)
\boldsymbol{\theta}^{T}\boldsymbol{x}_{a},
\end{equation}
\textit{where $C$ represents the normalizing 
	factor which is independent 
	of the unknown model parameters 
	$\boldsymbol{\theta}, 
	\alpha, \boldsymbol{\sigma}$.}
 
Drawing on the foundational lemma presented earlier, 
Algorithm \ref{Algo:1} introduces a method for 
posterior sampling tailored to address the challenges of 
the contextual bandit learning dilemma as discussed in Section 3. 
Each interaction cycle, or round $t$, commences with the identification 
of the model's parameters $\bm{\Psi}$, 
grounded in the posterior distribution outlined in Eq.(\ref{eque_posterior}). 
Subsequently, the algorithm computes the expected reward for 
each viable action, with a preference for 
selecting the action projected to offer the maximum return. 
Upon the decision maker's implementation of the chosen action, 
the system garners feedback $V_t(A_t)$ from the agent. 
This feedback is subsequently integrated into the decision history, 
thereby facilitating the transition to the subsequent iteration. 

\begin{algorithm}[htb]
\caption{TS-Conf (Thompson Sampling under Conformity)}
\label{Algo:1}
\begin{algorithmic}[1]
\STATE Initialize $\mathcal{H}_0 = \emptyset$
\FOR{$t = 1, 2, 3, \ldots, T$}
\STATE $\bm{\Psi}_t 
\sim p(\bm{\Psi}|
\mathcal{H}_{t-1})$ derived in Eq.          (\ref{eque_posterior})
\label{thm:Alg1:Initialize}
\STATE Select action by $A_t 
\leftarrow \arg
\max_{a\in \mathcal{A}_t} 
\mathbb{E}[{R}_{t}(a; \bm{\Psi}_t)]$ 
\STATE Observe the user feedback 
$V_t(A_t)$
\STATE Update history $\mathcal{H}_t 
\leftarrow 
\mathcal{H}_{t-1} \cup 
[A_t,h_{t,{A_t}},\boldsymbol{x}_{A_t},
V_t(A_t)]$
\ENDFOR
\end{algorithmic}
\end{algorithm}

In general, the posterior distribution derived in Eq. (\ref{eque_posterior}) 
is computationally expensive to sample exactly.  
Algorithm \ref{Algo:2} outlines the design of the TS-ConfMCMC algorithm, 
which applies the Markov Chain 
Monte Carlo (MCMC) technique 
to sample the posterior approximately.  
Specifically, it is a three-stage Gibbs sampler.
It is important to note 
that Eq.(\ref{eque_posterior}) 
implies that the conditional posterior distribution of ${\sigma}_a$ 
is independent across 
different actions $a$, given $\boldsymbol{\theta}$, $\alpha$. 
The preference vector 
$\boldsymbol{\theta}$ and 
the conformity tendency $\alpha$
manifest linearly when two other parameters are provided. 
We design a three-stage Gibbs sampler 
to efficiently facilitate the sampling process delineated in 
step \ref{thm:Alg1:Initialize} of the TS-Conf algorithm. 

\begin{algorithm}[htb]
\caption{TS-ConfMCMC}
\label{Algo:2}
\begin{algorithmic}[1]
\STATE Initialize $\mathcal{H}_0 = \emptyset$; 
$\sigma_{a,0}$;
$\boldsymbol{\theta}_{0}$;
${\alpha}_{0}$
\FOR{$t = 1, 2, 3, \ldots, T$}
\label{thm:Alg2:Initialize}
\FOR{$n = 1, 2, 3, \ldots, N$} 
\STATE $\boldsymbol{\theta}^{(n)} 
\sim p(\boldsymbol{\theta}|
\alpha = \alpha^{(n-1)}, 
\boldsymbol{\sigma} = \boldsymbol{\sigma}^{(n-1)}, 
\mathcal{H}_{t-1})$ 
\STATE $\alpha^{(n)} 
\sim p(\alpha|
\boldsymbol{\sigma}^{(n-1)}, 
\boldsymbol{\theta} = 
\boldsymbol{\theta}^{(n)}, 
\mathcal{H}_{t-1})$
\STATE $\boldsymbol{\sigma}_a^{(n)} 
\sim p(\boldsymbol{\sigma}_a|\alpha = 
\alpha^{(n)}, \boldsymbol{\theta} = 
\boldsymbol{\theta}^{(n)}, 
\mathcal{H}_{t-1}), \forall a$
\ENDFOR
\label{thm:Alg2:EndSample}
\STATE $\bm{\Psi}_t 
\leftarrow (\boldsymbol {\theta}^{(N)}, 
\alpha^{(N)}, 
\boldsymbol{\sigma}^{(N)} )$
\STATE Step 4-6 of Algorithm \ref{Algo:1}.

\ENDFOR
\end{algorithmic}
\end{algorithm}


\subsection{Regret Analysis}

Given a parameter $\bm{\Psi}$, denote the estimator 
for estimating $\alpha$ from $\mathcal{H}_t$ as $\widehat{\alpha}_t (\bm{\Psi})$.  
Note that this estimator is defined to assist the proof of regret, 
we do not need to know how to construct it.  
We define the confidence bound  $\widehat{\alpha}_t (\bm{\Psi})$ as: 
\[
\mathbb{P}
[\forall t,  
|\widehat{\alpha}_t (\bm{\Psi}) - \alpha| 
\leq 
W_t(\delta; \bm{\Psi})
]
\geq 
1- \delta.
\]
Different instances of $\widehat{\alpha}_t (\bm{\Psi})$ have 
different confidence width.  
Let $W^\ast_t (\delta; \bm{\Psi})$ denote the smallest possible confidence width attained 
by the optimal estimator $\widehat{\alpha}^\ast_t (\bm{\Psi})$.   


\begin{theorem}
\label{RegretUpper}
The regret of Algorithm 1 satisfies: 
\begin{align*}
R_T(\mathcal{D})  
&\leq 
O\left( 
\frac{1}{1-\alpha} 
\int 
\sum\nolimits^T_{t=1} W^\ast_t (1/T; \bm{\Psi}) d \bm{\Psi} \right.  + \left. \frac{1}{1-\alpha} d \sqrt{T} \ln T 
\right).
\end{align*}
\end{theorem}

The following theorem states that the above bound is tight and 
reveals that our algorithm is guaranteed to converge at a sublinear rate. 
\begin{theorem}
\label{RegetLower}
If $\sum^T_{t=1} W^\ast_t (1/T; \bm{\Psi})  = \Omega(T)$ holds for all $\bm{\Psi}$, 
the regret is lower bounded by:   
\[
R_T(\mathcal{D}) 
\geq
\Omega(T).
\]  
\end{theorem}

The above theorem shows that the algorithm can effectively 
learn from historical decisions and gradually approach 
the performance of the best possible decision. \textit{Due to page limitations, more details on the proofs of the above two theorems can be found in our technical report \cite{xu2024}.} 

\section{Experiments}
\subsection{Experiment Settings}
\noindent \textbf{Datasets.}  
Our approach to constructing the simulated datasets is aligned with established practices in the field, similar to those employed in related studies\cite{xu2021generalized,mo2023multi}. 
In our empirical evaluation using 
real-world data, we employ datasets sourced from 
four distinct platforms: 
Amazon Music\footnote{\url{http://jmcauley.ucsd.edu/data/amazon/index\_2014.html}}, 
MovieLens\footnote{\url{https://grouplens.org/datasets/movielens/}}, 
Yelp\footnote{\url{https://www.yelp.com/dataset}}, 
and Google Maps\footnote{\url{http://jmcauley.ucsd.edu/data/googlelocal/}}. 
Through the utilization of these varied 
datasets spanning multiple domains, 
our objective is to rigorously assess and 
understand the real-world applicability and 
performance of the proposed algorithm. 
Based on the approach similar to \cite{xu2021generalized,zhao2022knowledge}, 
we perform data preprocessing and assess the accuracy of 
algorithm recommendations through regret values. 
\textit{Due to page limitations, more details on 
the data preprocessing can be found in our technical report \cite{xu2024}.}

\noindent \textbf{Baselines and metrics.}
To the best of our knowledge, 
limited research exists on 
contextual bandit algorithms that 
specifically address the herding effects. 
In light of this gap, 
we adapt mainstream bandit algorithms to 
incorporate the herding effects, 
thereby producing a suitable comparison algorithm. 
To ensure a fair and relevant comparison in our study, 
which focuses on the herding effects, 
we benchmark our proposed algorithm against 
two sets of algorithms. 
The first set includes established baseline algorithms, 
LinUCB\cite{chu2011contextual} and 
Thompson Sampling (TS)\cite{agrawal2013thompson}, 
known for their accuracy in scenarios 
with unbiased user feedback. 
However, these algorithms do not specifically 
address herding effects, a gap in the existing research. 
Recognizing this limitation, we developed LinUCBConf, 
an adaptation of the LinUCB algorithm. 
LinUCBConf is designed to provide a more 
appropriate baseline for our study by accounting 
for herding effects, which were not explicitly 
considered in previous models. 
This adaptation allows for a more equitable comparison, 
enabling us to effectively demonstrate the strengths 
and innovations of our proposed algorithm in 
the context of herding effects. 
When estimating the 
preference parameter $\boldsymbol{\theta}_t$, 
LinUCBConf employs the same action selection method 
as LinUCB, using $\widetilde{\boldsymbol{x}}_{A(\tau)}$ as 
the feature vector and $V_t(A_t)$ for feedback 
to learn user preferences $\boldsymbol{\theta}$.
To evaluate the efficacy of each algorithm,
our primary metric is the regret value, 
as delineated in Eq.\ref{equa4}. 

\subsection{Stability Analysis of Algorithm Parameters}
\begin{figure}
	\centering
	\begin{subfigure}[b]{0.24\textwidth}
		\includegraphics[width=\textwidth]{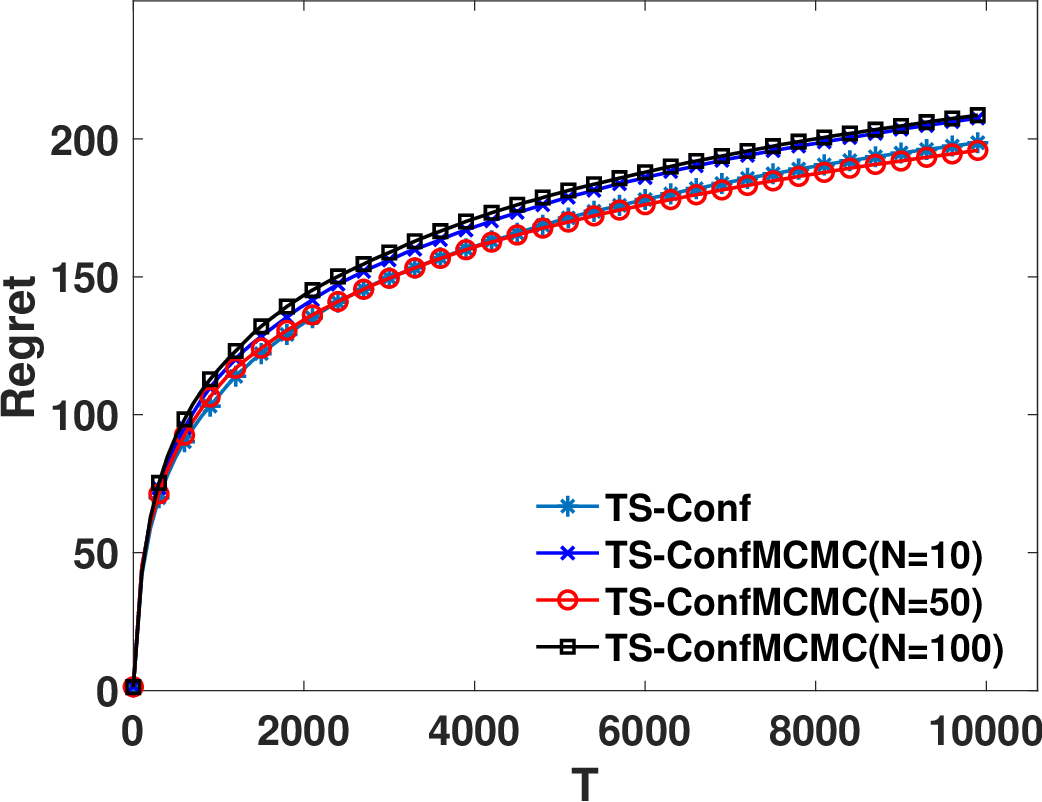}
		\subcaption{Dimension $d=5$}
	\end{subfigure}
	\begin{subfigure}[b]{0.24\textwidth}
		\includegraphics[width=\textwidth]{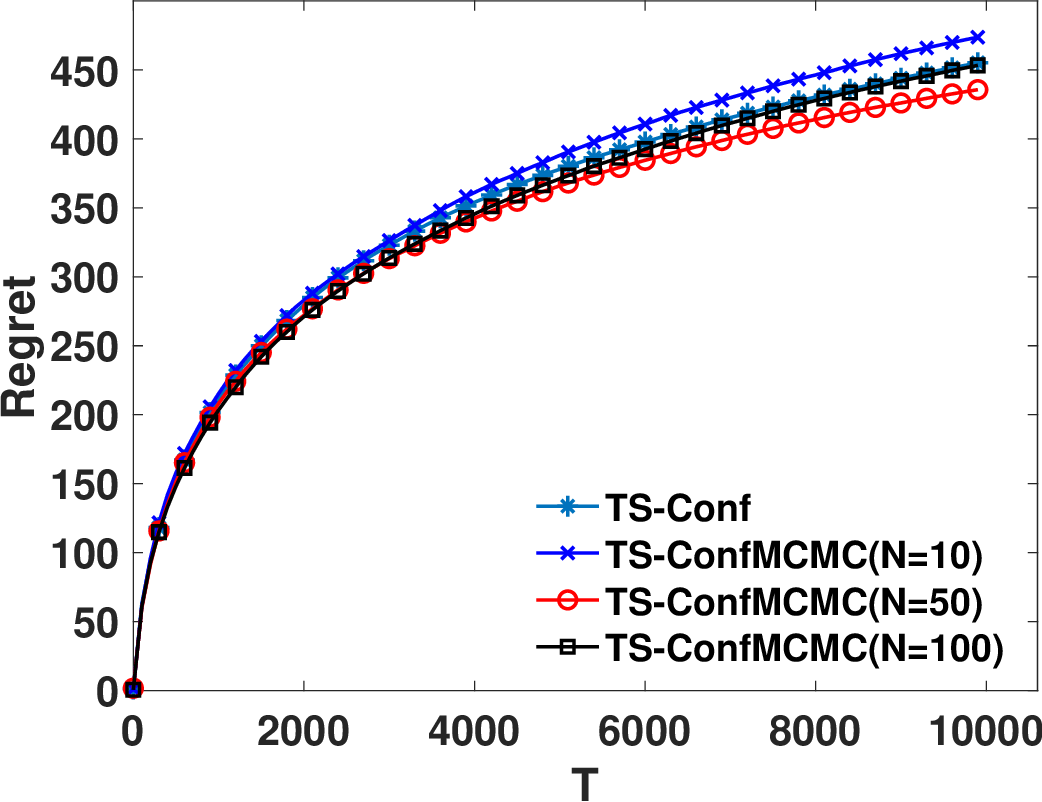}
		\subcaption{Dimension $d=10$}
	\end{subfigure}
	\begin{subfigure}[b]{0.24\textwidth}
		\includegraphics[width=\textwidth]{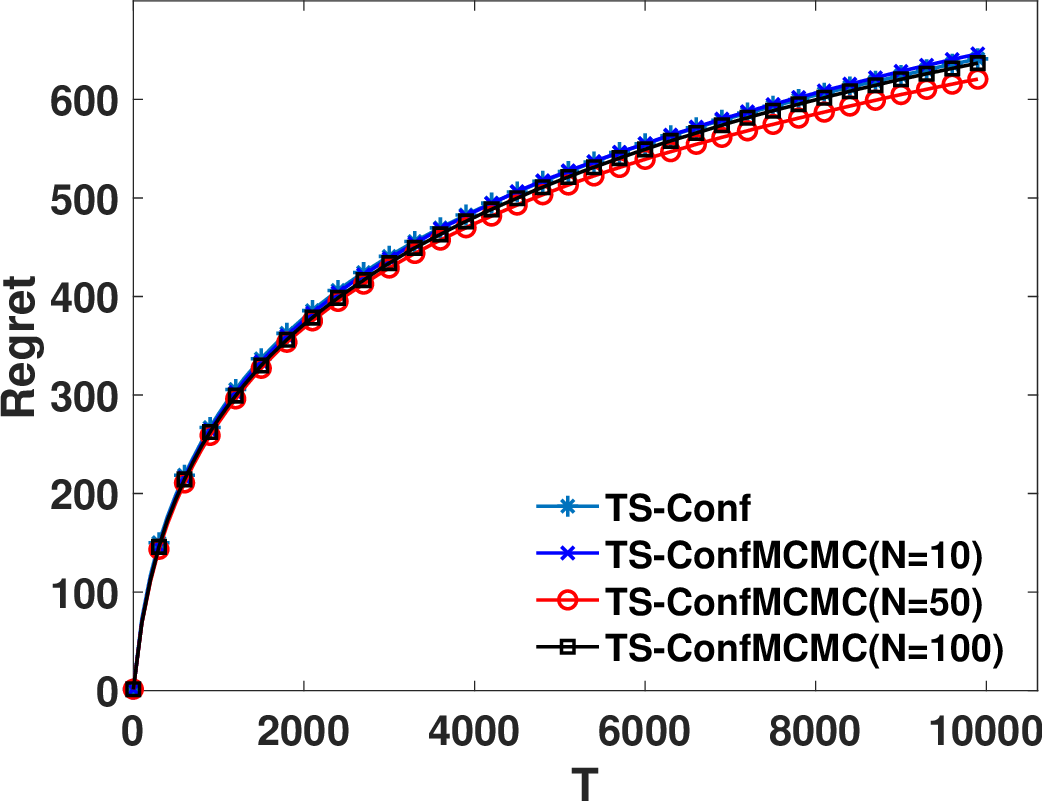}
		\subcaption{Dimension $d=15$}
	\end{subfigure}
	\begin{subfigure}[b]{0.24\textwidth}
		\includegraphics[width=0.99\textwidth]{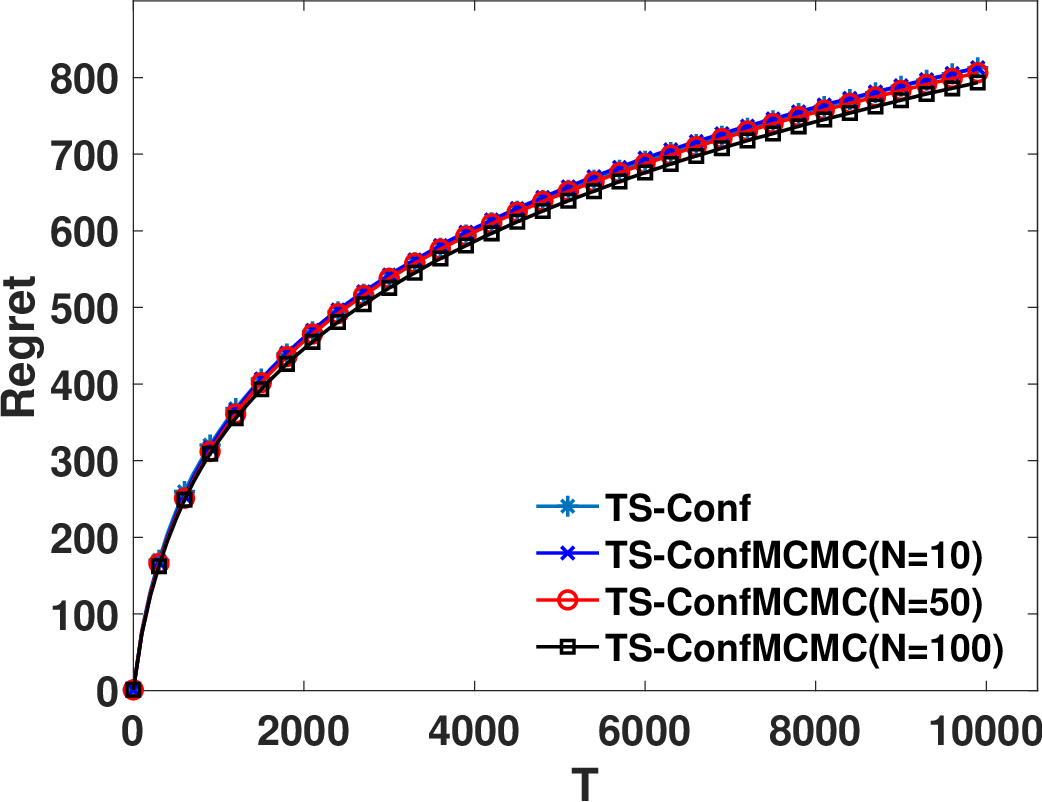}
		\subcaption{Dimension $d=20$}
	\end{subfigure}
	\centering
	\begin{subfigure}[b]{0.24\textwidth}
		\includegraphics[width=\textwidth]{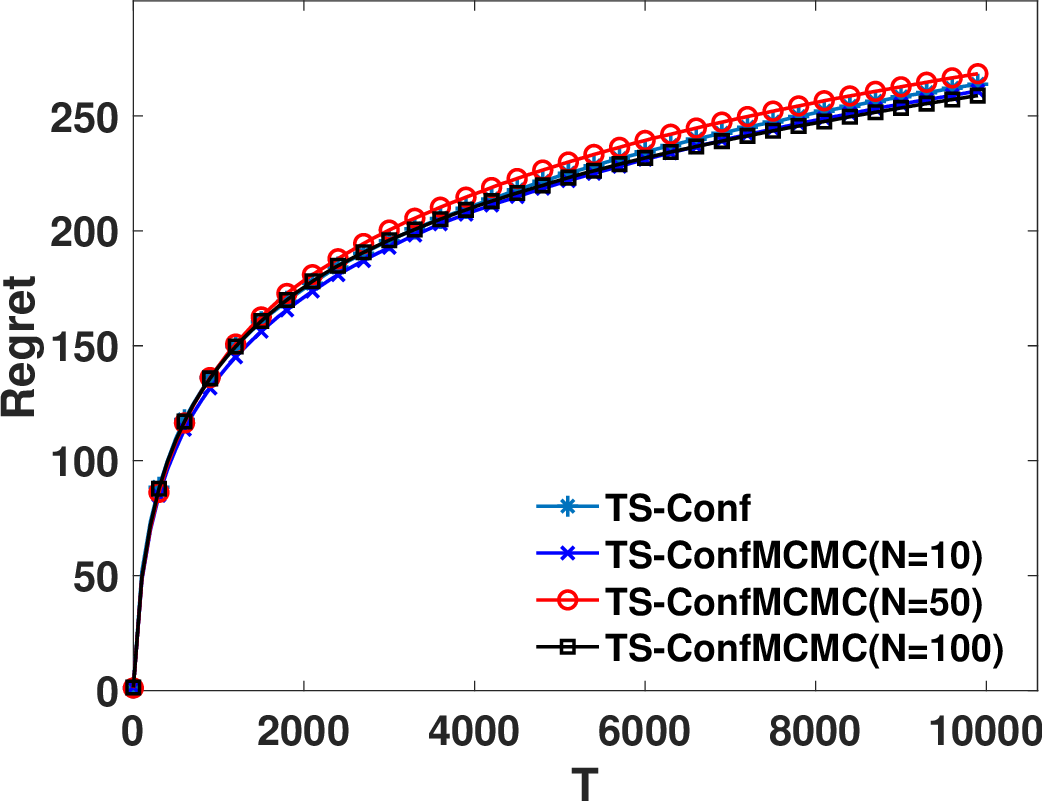}
		\subcaption{Noise Var $\sigma^2=0.5$}
	\end{subfigure}
	\begin{subfigure}[b]{0.24\textwidth}
		\includegraphics[width=\textwidth]{pic/Alg1_noise_in/simulation/para_contrast/mcmc/dimension_10_noise_1.0.eps}
		\subcaption{Noise Var $\sigma^2=1.0$}
	\end{subfigure}
	\begin{subfigure}[b]{0.24\textwidth}
		\includegraphics[width=\textwidth]{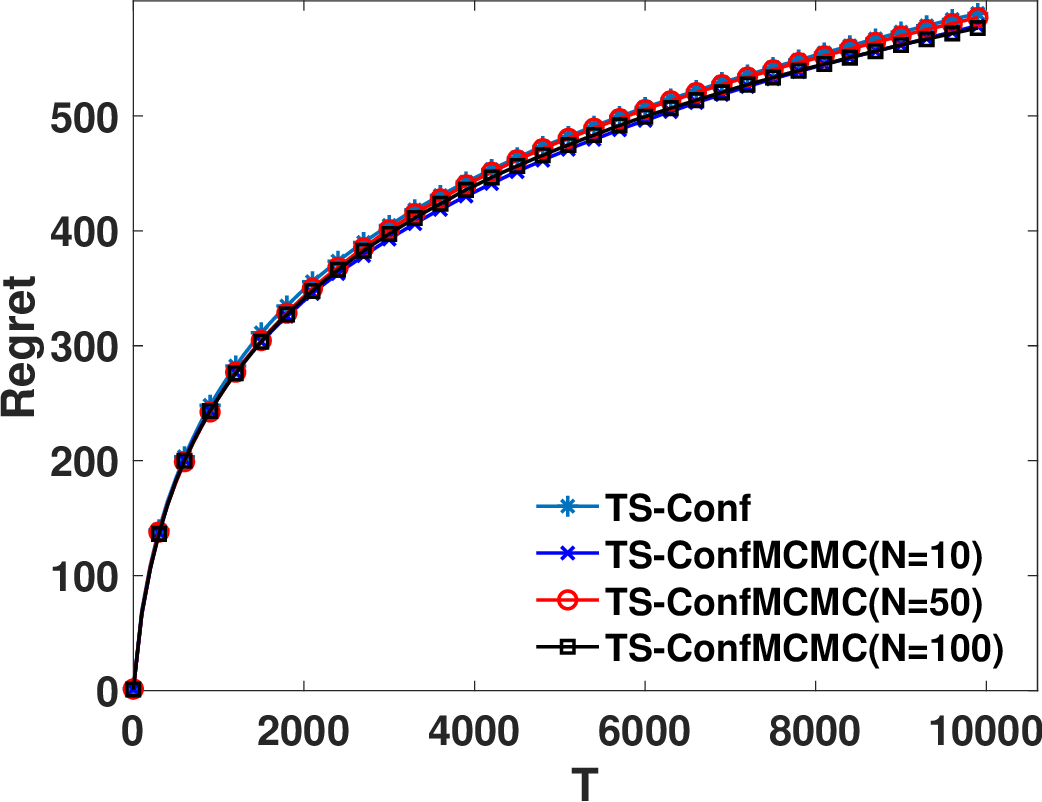}
		\subcaption{Noise Var $\sigma^2=1.5$}
	\end{subfigure}
	\begin{subfigure}[b]{0.24\textwidth}
		\includegraphics[width=0.99\textwidth]{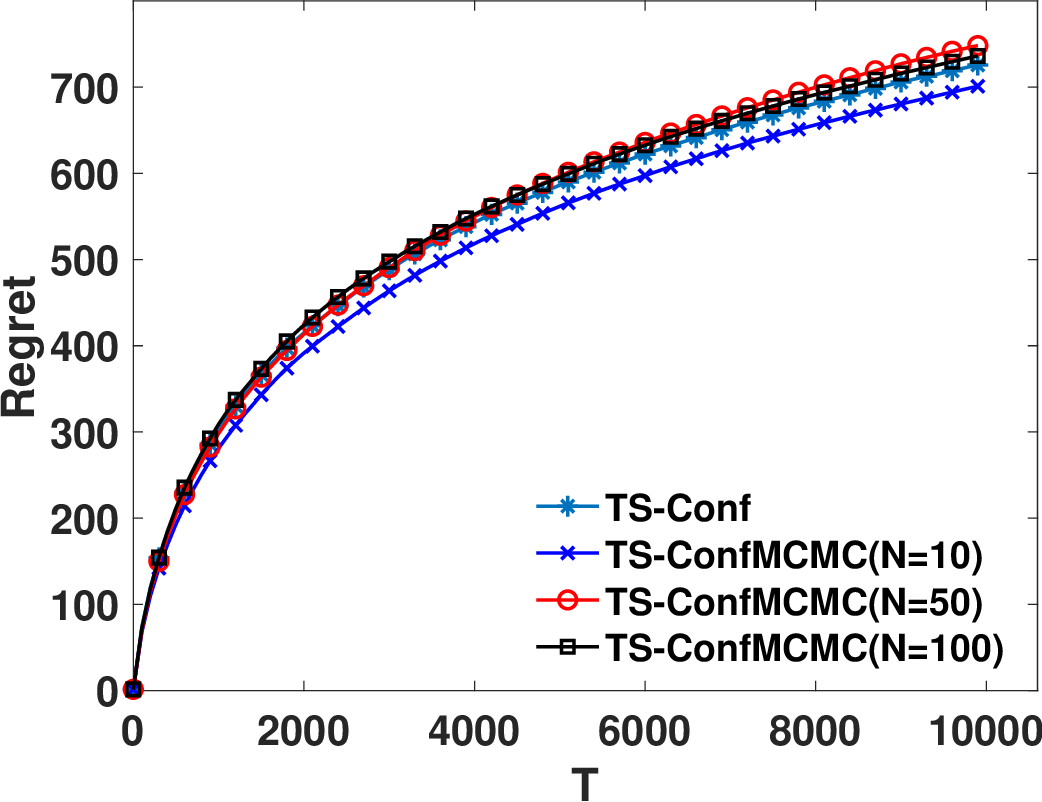}
		\subcaption{Noise Var $\sigma^2=2.0$}
	\end{subfigure}
	\caption{Comparative analysis of TS-Conf 
		and TS-ConfMCMC 
  } 
	\label{figmcmc}
\end{figure}

\noindent \textbf{Impact of the MCMC Approximation.} 
In our study, we focus on analyzing the influence of 
the number of iterations, denoted as $N$, 
on the regret metric of our algorithm. 
The parameter $N$ is critical as it represents 
the iterations necessary for the MCMC method to 
approximate samples for the posterior distribution. 
For the purpose of this analysis, 
we standardize the dimensions of the actions 
and the noise variance at $d = 10$ 
and $\sigma^2 = 1.0$, respectively. 
This uniformity allows for a controlled assessment of 
the impact of $N$ on the algorithm's regret. 
As illustrated in Figure \ref{figmcmc}, 
the regret values produced by 
TS-Conf consistently fall within the 
range spanned by TS-ConfMCMC. 
This shows that the approximate algorithm TS-ConfMCMC 
we proposed can obtain results similar to precise sampling 
through a limited number of samplings, illustrating the 
effectiveness of the approximate algorithm.


\begin{figure}
\centering
\begin{subfigure}[b]{0.24\textwidth}
\includegraphics[width=\textwidth]{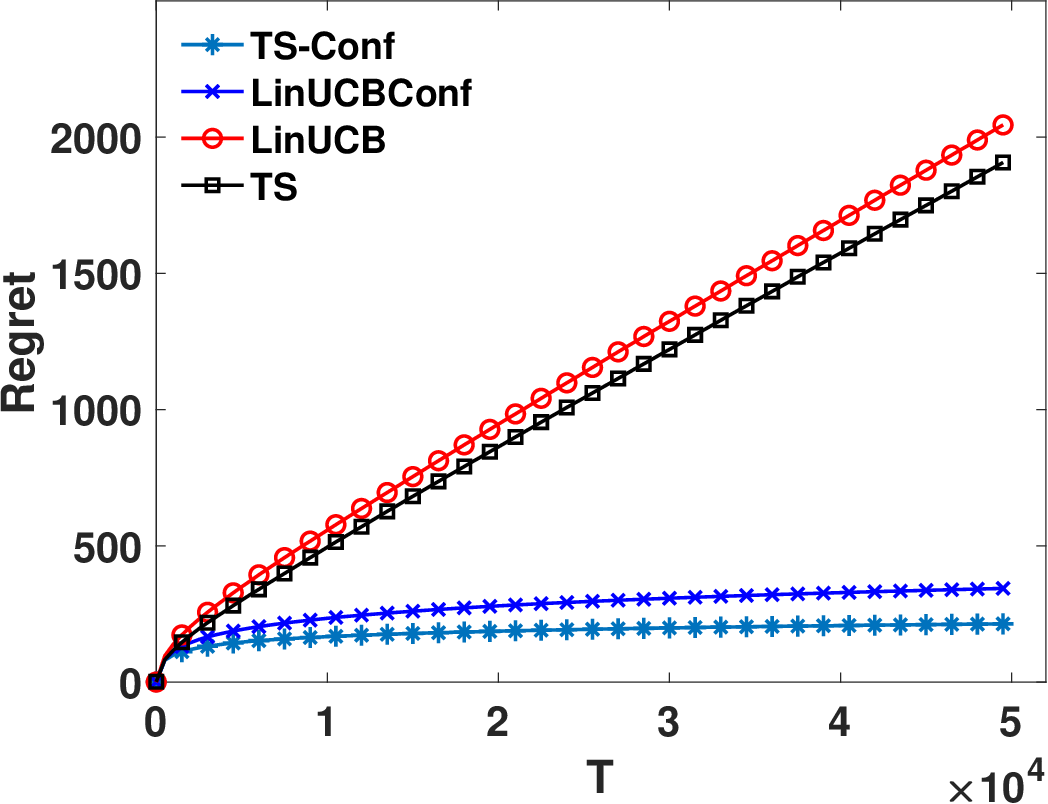}
\subcaption{Dimension $d=5$}
\label{fig1(a)}
\end{subfigure}
\begin{subfigure}[b]{0.24\textwidth}
\includegraphics[width=\textwidth]{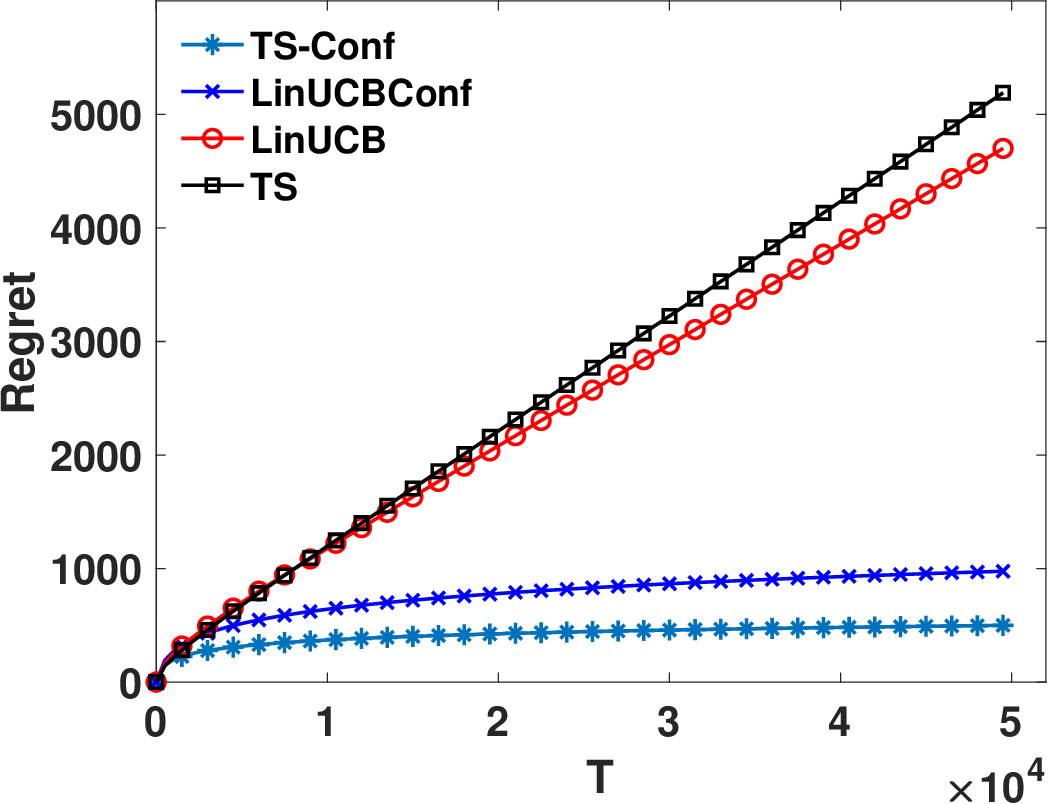}
\subcaption{Dimension $d=10$}
\label{fig1(b)}
\end{subfigure}
\begin{subfigure}[b]{0.24\textwidth}
\includegraphics[width=\textwidth]{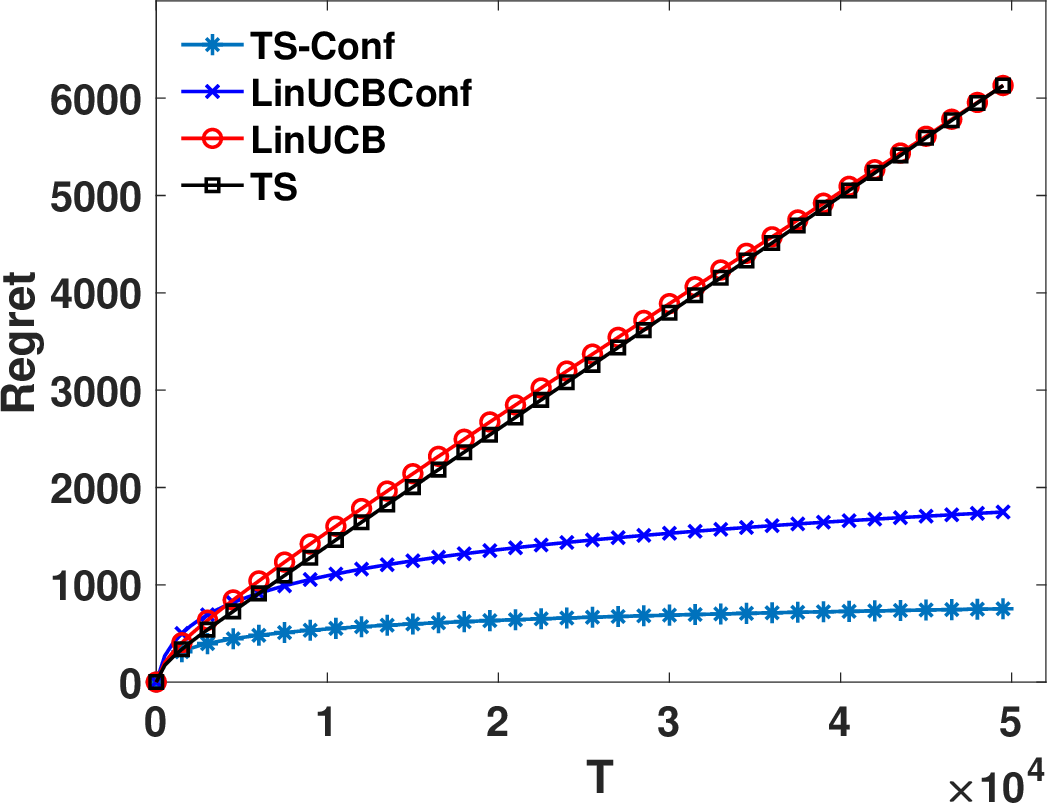}
\subcaption{Dimension $d=15$}
\label{fig1(c)}
\end{subfigure}
\begin{subfigure}[b]{0.24\textwidth}
\includegraphics[width=0.99\textwidth]{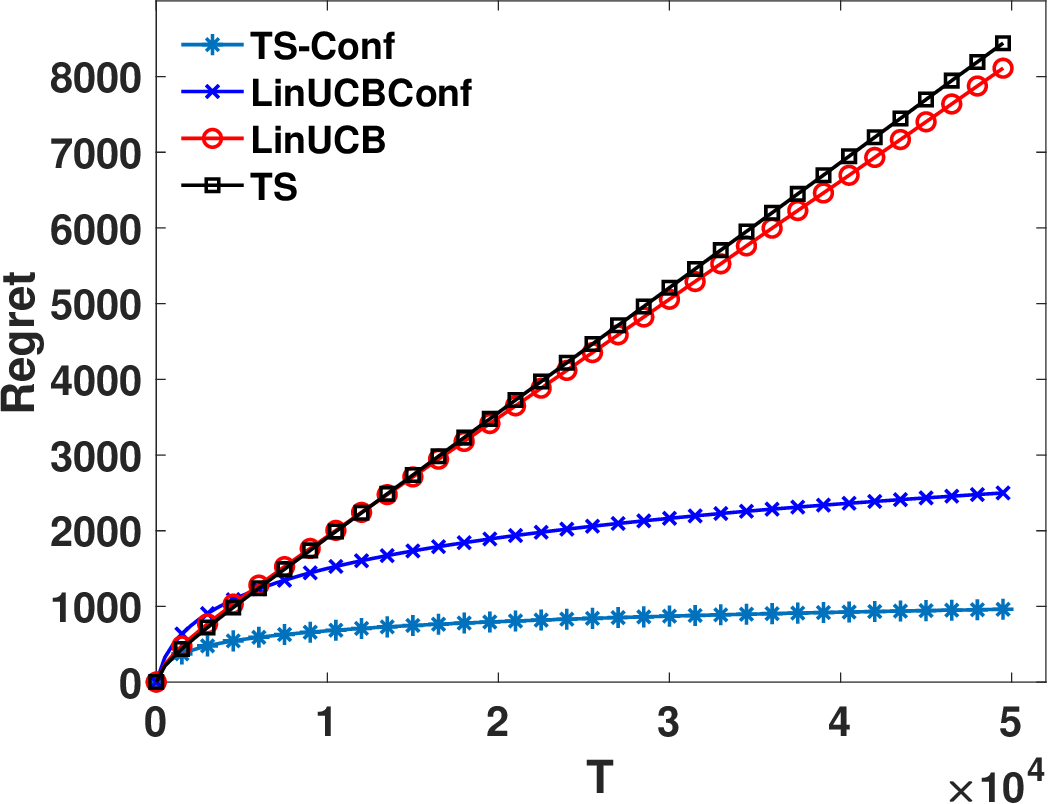}
\subcaption{Dimension $d=20$}
\label{fig1(d)}
\end{subfigure}
\caption{Impact of dimensions $d$ in synthetic dataset} 
\label{fig1}
\end{figure}

\noindent \textbf{Impact of dimensions $d$.}
This section delves into the effect 
of varying dimensionality on the regret value 
$\hat{R}_{t}$ across different algorithms. 
We keep the constant noise variance $\sigma^2=1.0$ for 
consistency in our experiments. 
We evaluate the performance under four dimensions: 
$d=5$, $d=10$, $d=15$, and $d=20$.
Figure \ref{fig1(a)} shows that 
when the feature dimension is $d=5$, 
the regret curve for TS-Conf is 
the lowest among the four algorithms. 
It suggests that TS-Conf consistently yields 
lower cumulative regret values than 
the other three baselines. 
The LinUCB and TS algorithms exhibit significantly 
higher regret values than TS-Conf. 
This observation underscores the pitfalls of 
assuming unbiased user feedback in the presence 
of the herding effects, highlighting the necessity 
to address such biases. 
While LinUCBConf does account for 
biased user feedback, its regret values remain 
higher than those of TS-Conf. 
This difference underscores the varying outcomes 
that different exploration-exploitation 
trade-off strategies can produce. 
It attests to the efficacy of Bayesian-based 
posterior sampling techniques in managing uncertainty.
As the feature dimension increases to $d=10$, 
$d=15$, and $d=20$, 
Figures \ref{fig1(b)}, \ref{fig1(c)}, and 
\ref{fig1(d)} mirror the above trends, 
indicating that these observations persist across 
different feature dimensions. 

\begin{figure}
\centering
\begin{subfigure}[b]{0.24\textwidth}
\includegraphics[width=\textwidth]{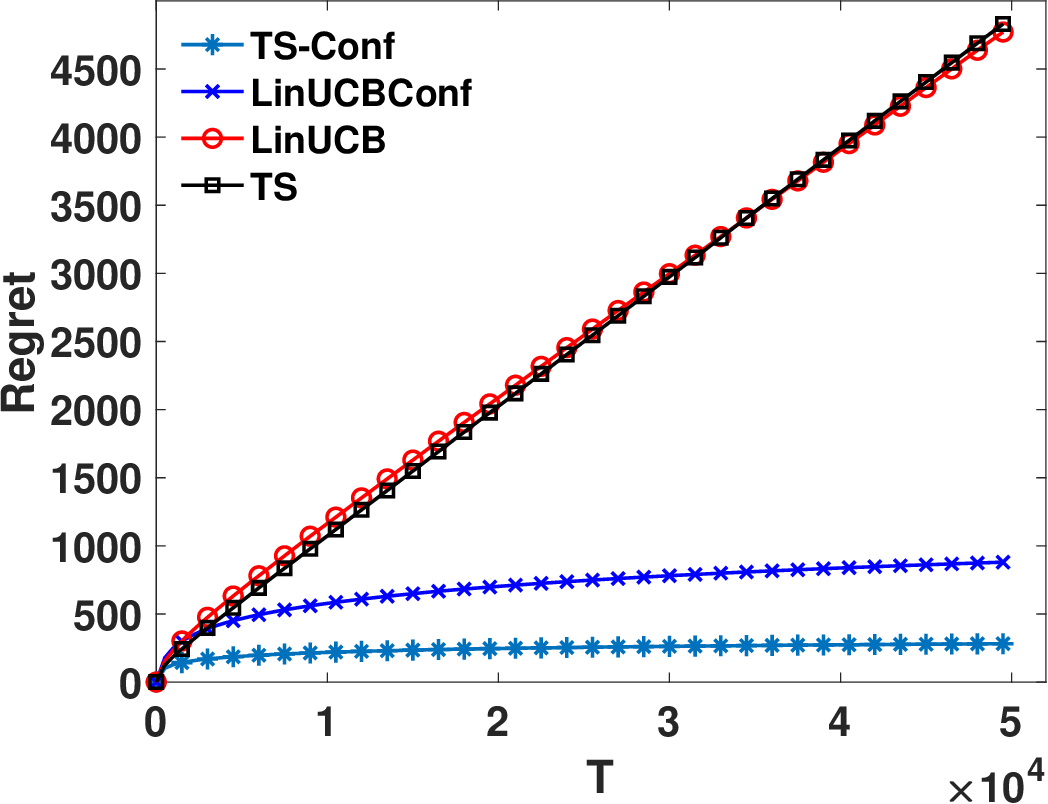}
\subcaption{Noise Var $\sigma^2=0.5$}
\label{fig2(a)}
\end{subfigure}
\begin{subfigure}[b]{0.24\textwidth}
\includegraphics[width=\textwidth]{pic/Alg1_noise_in/simulation/algo_contrast/regret/dimension_10_noise_1.0.eps}
\subcaption{Noise Var $\sigma^2=1.0$}
\label{fig2(b)}
\end{subfigure}
\begin{subfigure}[b]{0.24\textwidth}
\includegraphics[width=\textwidth]{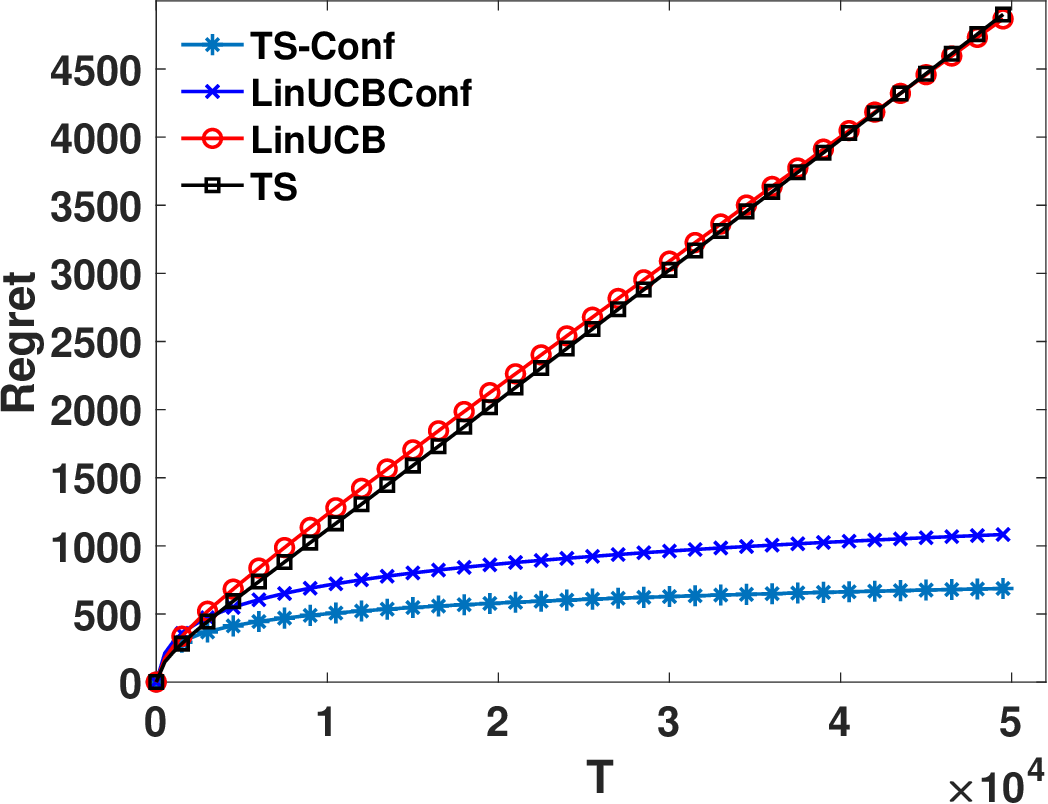}
\subcaption{Noise Var $\sigma^2=1.5$}
\label{fig2(c)}
\end{subfigure}
\begin{subfigure}[b]{0.24\textwidth}
\includegraphics[width=0.99\textwidth]{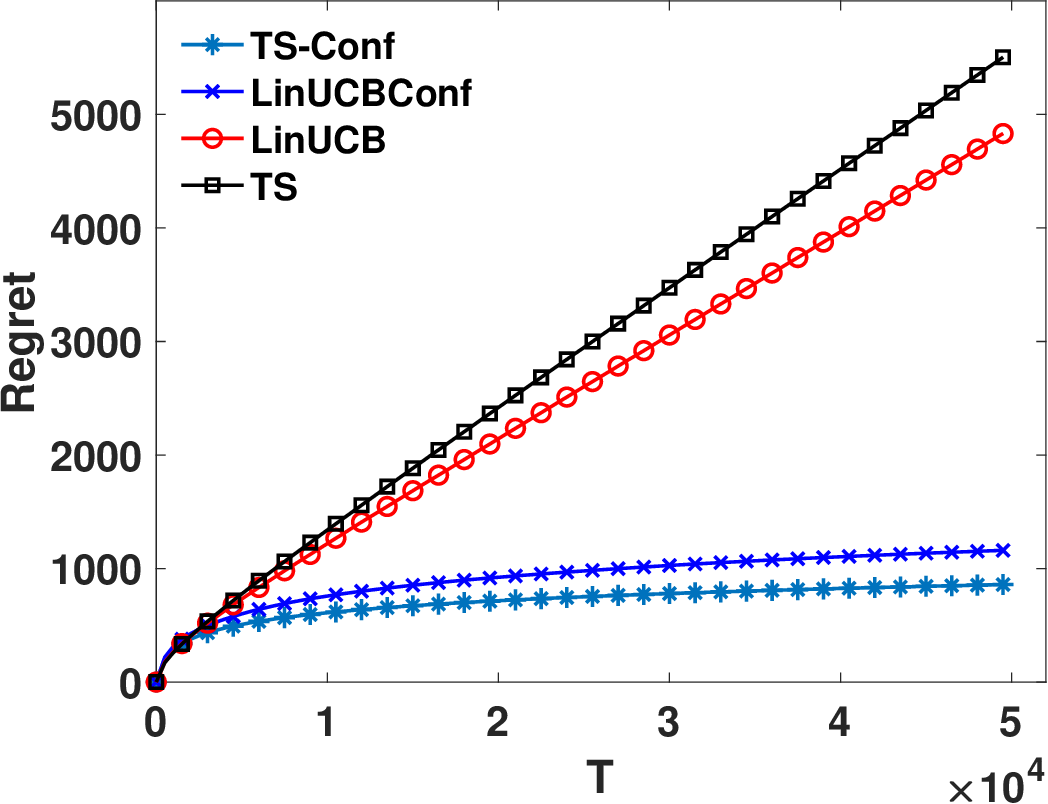}
\subcaption{Noise Var $\sigma^2=2.0$}
\label{fig2(d)}
\end{subfigure}
\caption{Impact of  
noise variance $\sigma^2$ in synthetic dataset} 
\label{fig2}
\end{figure}

\noindent \textbf{Impact of noise variance $\sigma^2$.} 
In this section, we delve into the effect 
of varying noise variance on the regret value 
$\hat{R}_{t}$ across different algorithms. 
Consistently, our experiments are conducted with 
a feature dimension $d=10$. We assess performance 
under four distinct noise variance settings: 
$\sigma^2=0.5$, $\sigma^2=1.0$, 
$\sigma^2=1.5$, and $\sigma^2=2.0$. 
Figure \ref{fig2(a)} shows that with the noise variance of $\sigma^2=0.5$, 
TS-Conf consistently exhibits the lowest regret 
values among the four algorithms. 
LinUCB and TS, which overlook the bias in 
user feedback, and LinUCBConf, which employs 
the LinUCB approach for exploration-exploitation 
trade-off, register significantly higher regret 
values than TS-Conf.
This trend persists as the noise variance increases 
to $\sigma^2=1.0$, $\sigma^2=1.5$, and 
$\sigma^2=2.0$, as evidenced in 
Figure \ref{fig2(b)}, Figure \ref{fig2(c)}, 
and Figure \ref{fig2(d)}. Such consistent 
performance under varying noise levels underscores 
the stability and robustness of the TS-Conf 
algorithm in handling uncertainties. 

\subsection{Real-world applications}
\begin{figure}
\centering
\begin{subfigure}[b]{0.24\textwidth}
\includegraphics[width=\textwidth]{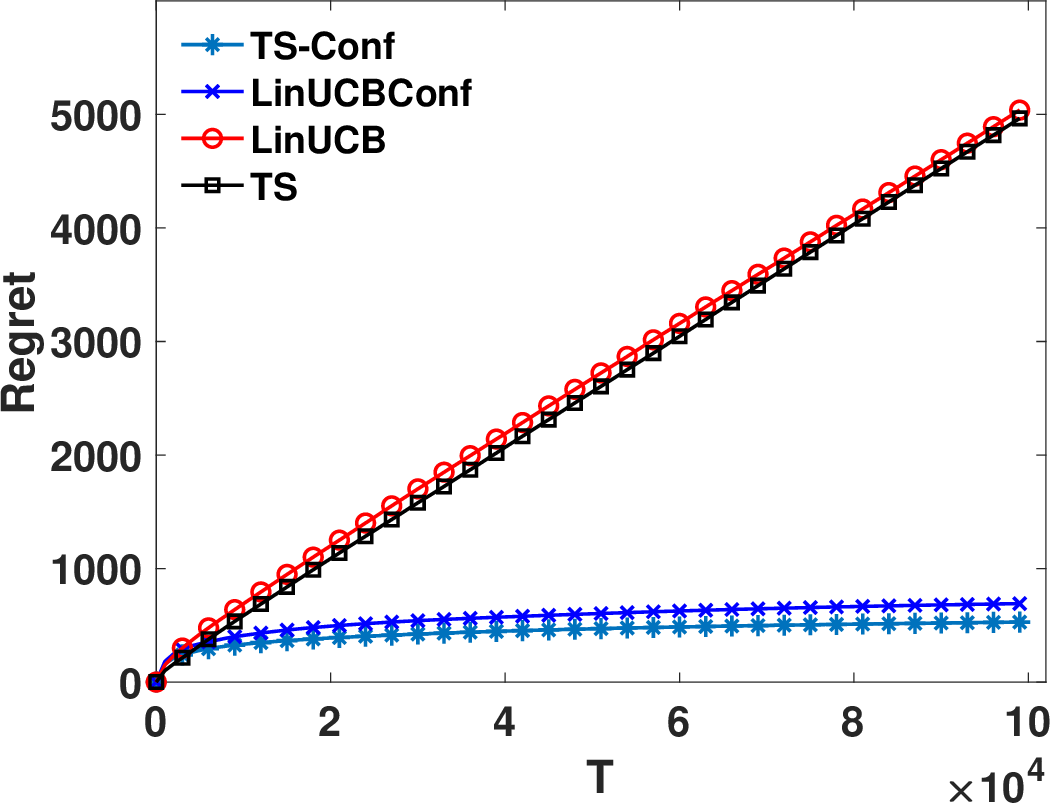}
\subcaption{Dimension $d=5$}
\end{subfigure}
\begin{subfigure}[b]{0.24\textwidth}
\includegraphics[width=\textwidth]{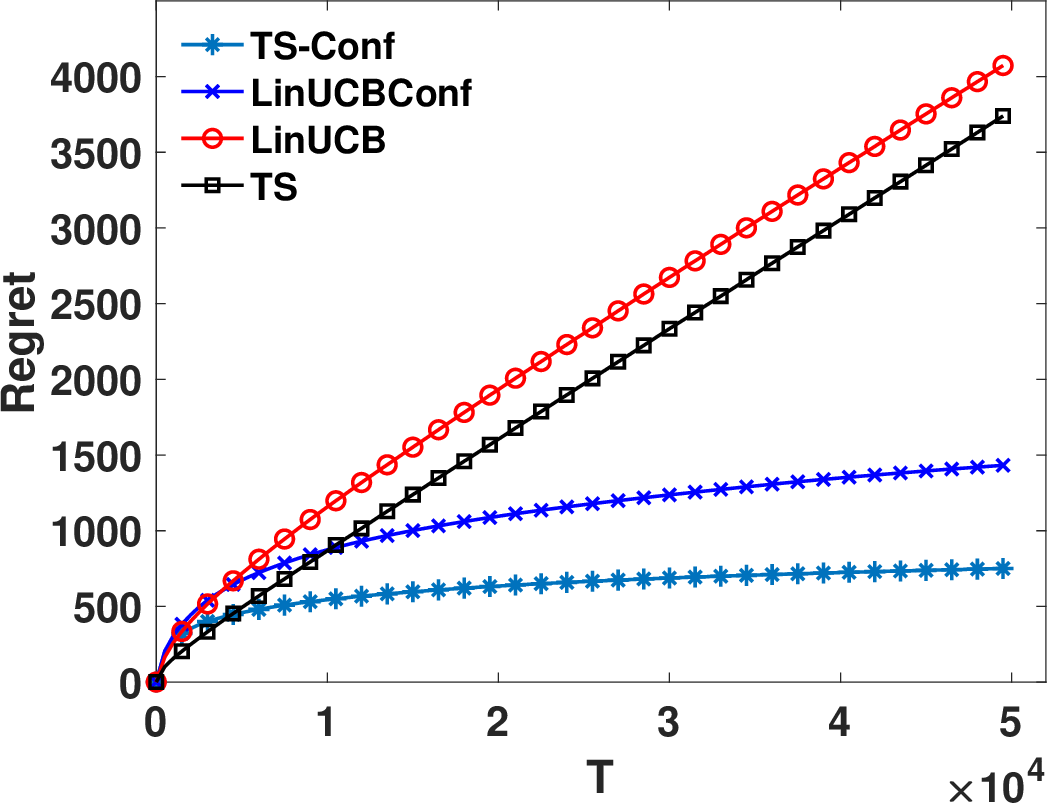}
\subcaption{Dimension $d=10$}
\end{subfigure}
\begin{subfigure}[b]{0.24\textwidth}
\includegraphics[width=\textwidth]{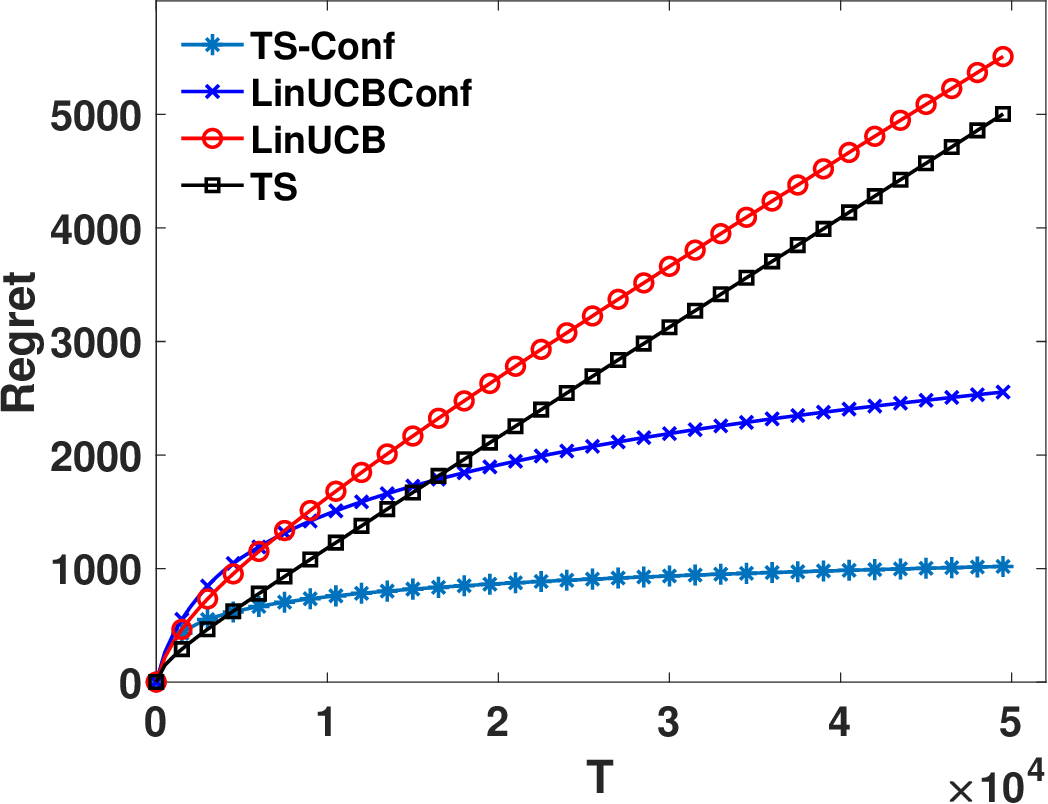}
\subcaption{Dimension $d=15$}
\end{subfigure}
\begin{subfigure}[b]{0.24\textwidth}
\includegraphics[width=0.99\textwidth]{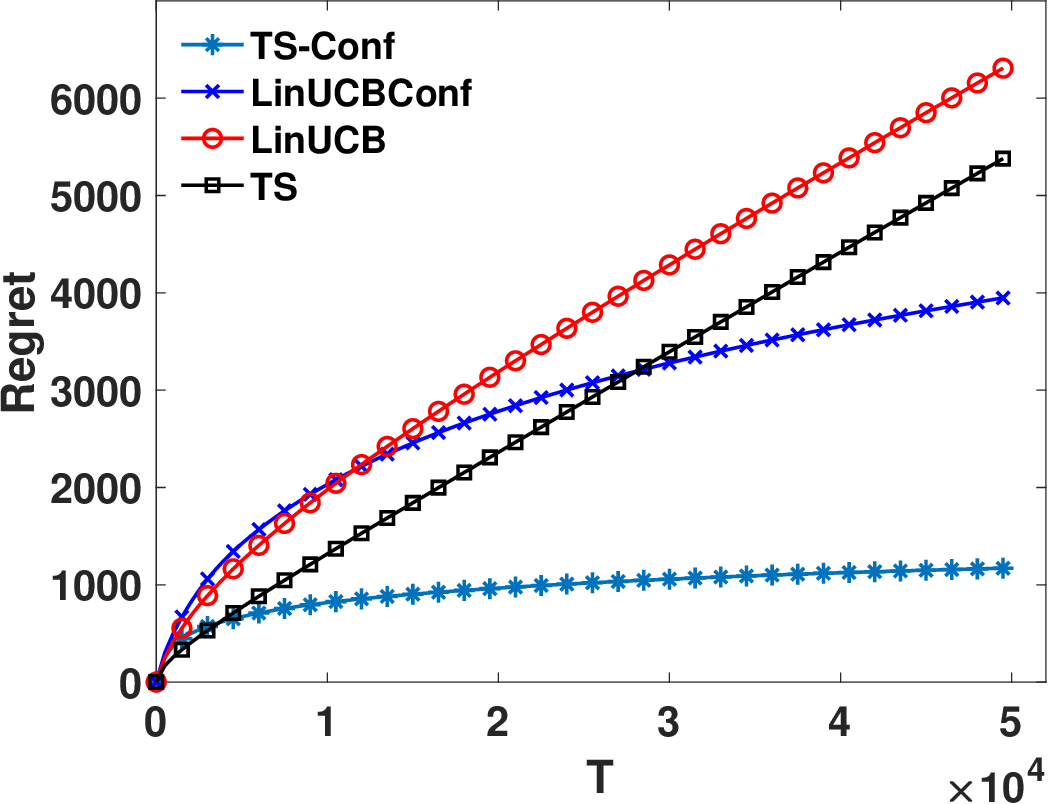}
\subcaption{Dimension $d=20$}
\end{subfigure}
\centering
\begin{subfigure}[b]{0.24\textwidth}
\includegraphics[width=\textwidth]{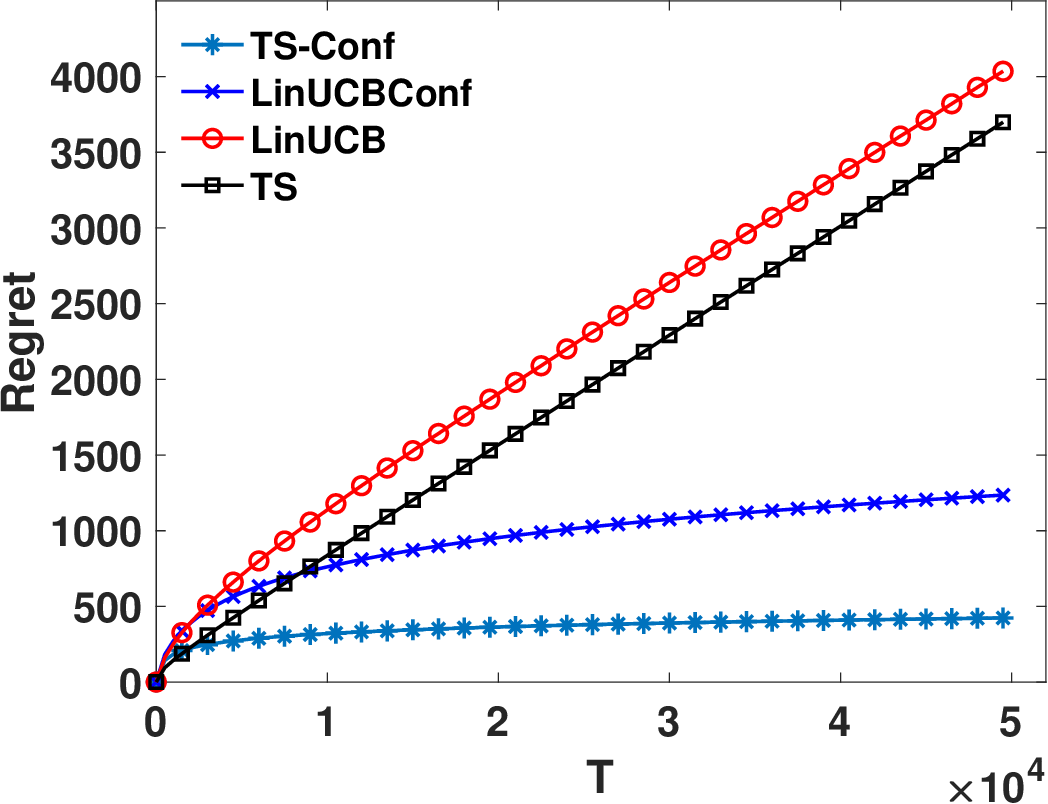}
\subcaption{Noise Var $\sigma^2=0.5$}
\end{subfigure}
\begin{subfigure}[b]{0.24\textwidth}
\includegraphics[width=\textwidth]{pic/Alg1_noise_in/MovieLens_1m/algo_contrast/regret/dimension_10_noise_1.0.eps}
\subcaption{Noise Var $\sigma^2=1.0$}
\end{subfigure}
\begin{subfigure}[b]{0.24\textwidth}
\includegraphics[width=\textwidth]{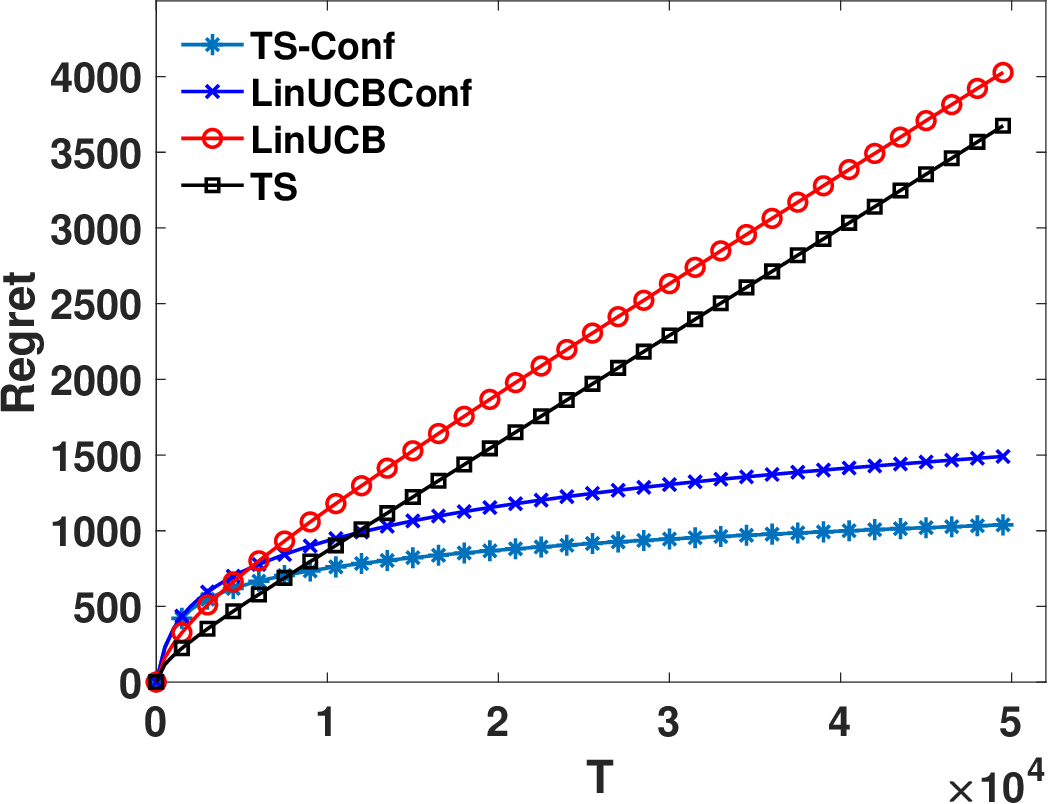}
\subcaption{Noise Var $\sigma^2=1.5$}
\end{subfigure}
\begin{subfigure}[b]{0.24\textwidth}
\includegraphics[width=0.99\textwidth]{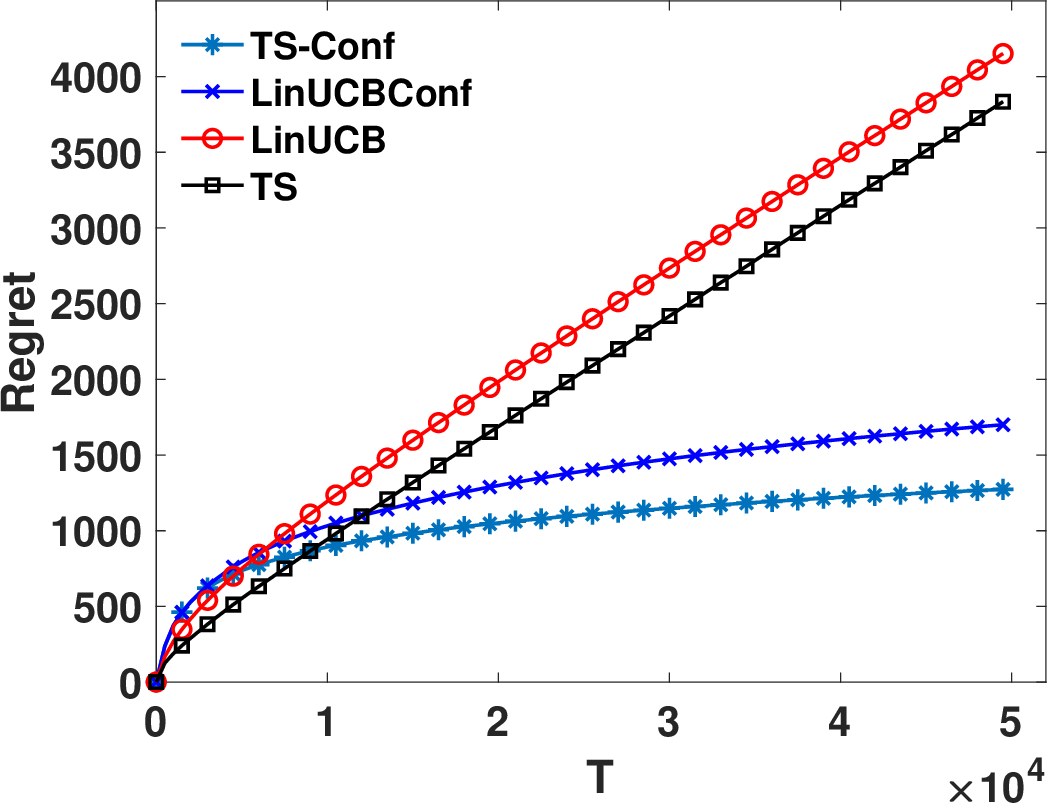}
\subcaption{Noise Var $\sigma^2=2.0$}
\end{subfigure}
\caption{Impact of dimensions $d$ and noise variance $\sigma^2$ in MovieLens dataset.
} 
\label{fig13}
\end{figure}

\noindent \textbf{Results in MovieLens.} 
Figure \ref{fig13} shows the regret $\hat{R}_{t}$
produced by each algorithm in different 
dimensions and different noise variances. 
It can be observed that 
the TS-Conf algorithm 
always has the lowest 
regret value 
across varying dimensions 
and noise levels. 
Consistent with the 
experimental observations 
on synthetic data,  
the regret values for 
both LinUCB and TS 
exhibit significantly 
higher regret than the 
TS-Conf algorithm and 
will increase linearly with $t$. 
Similarly, LinUCBConf 
also has a
greater regret compared with TS-Conf, 
and its regret values 
converge at a slower rate 
than our algorithm.

\begin{figure}
\centering
\begin{subfigure}[b]{0.24\textwidth}
\includegraphics[width=\textwidth]{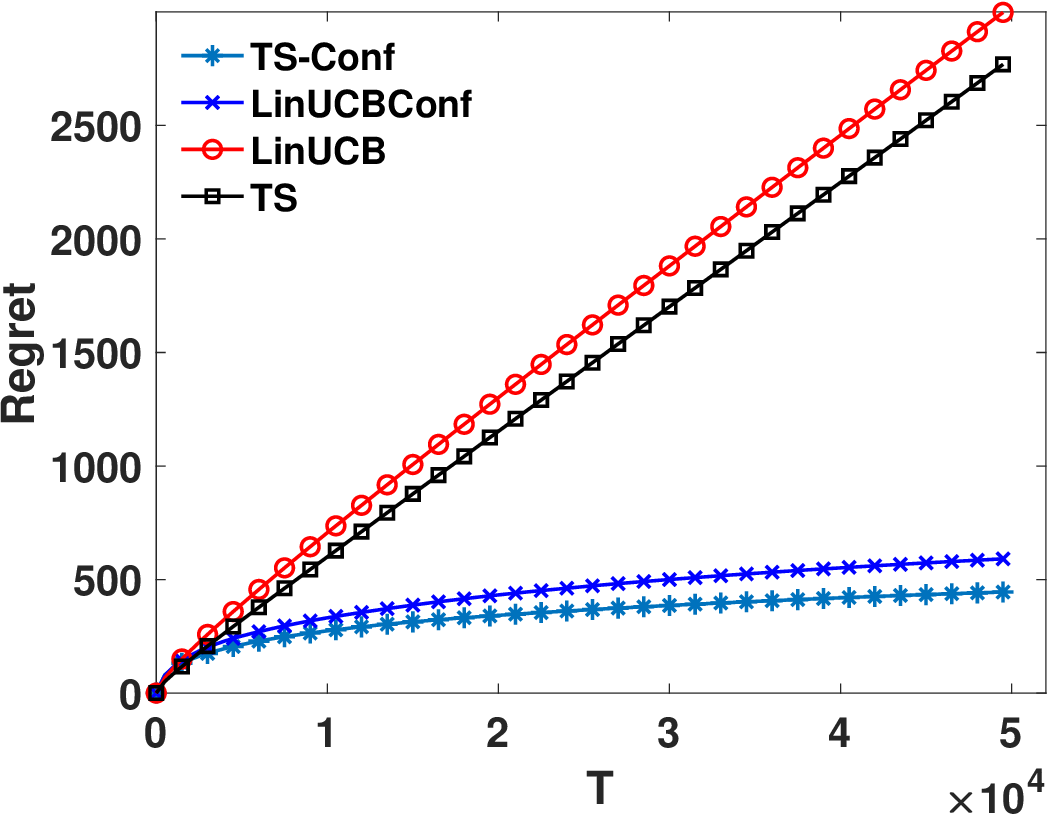}
\subcaption{Dimension $d=5$}
\end{subfigure}
\begin{subfigure}[b]{0.24\textwidth}
\includegraphics[width=\textwidth]{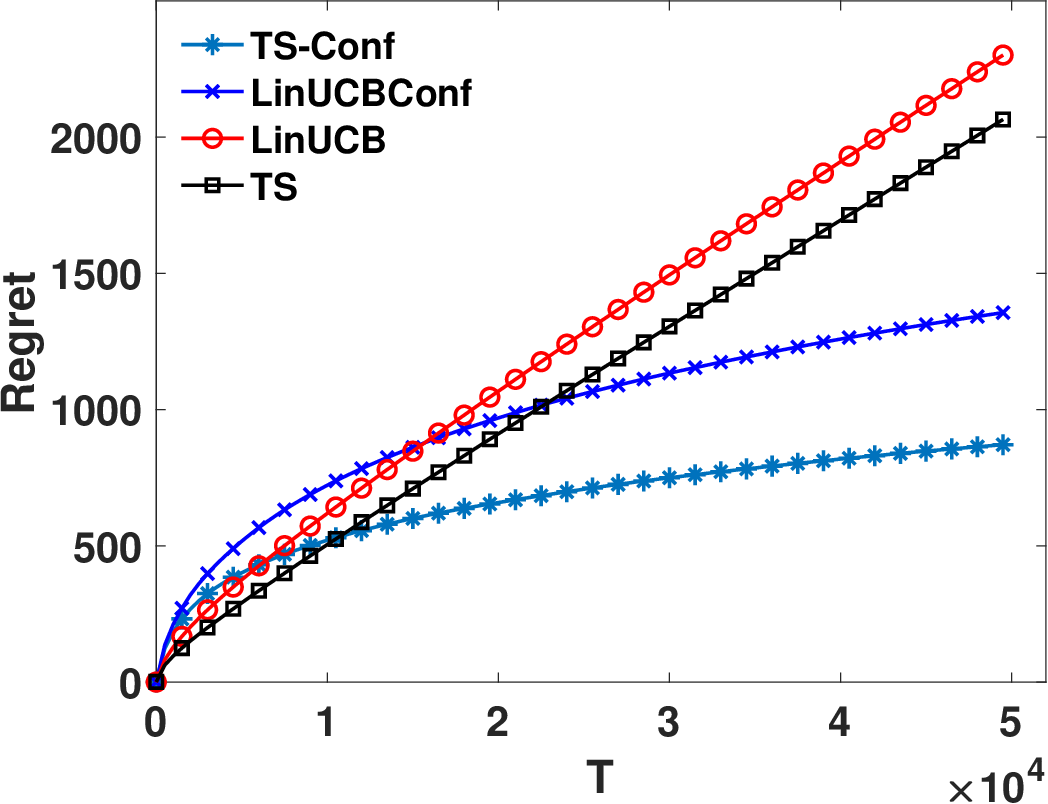}
\subcaption{Dimension $d=10$}
\end{subfigure}
\begin{subfigure}[b]{0.24\textwidth}
\includegraphics[width=\textwidth]{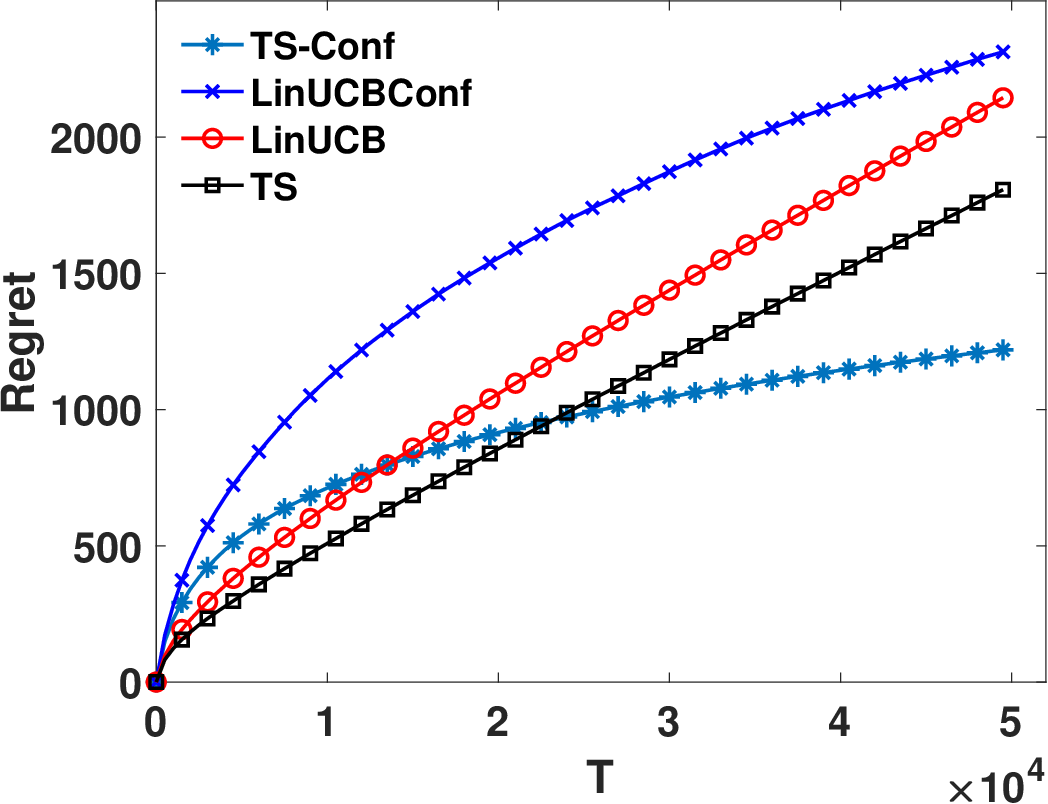}
\subcaption{Dimension $d=15$}
\end{subfigure}
\begin{subfigure}[b]{0.24\textwidth}
\includegraphics[width=0.99\textwidth]{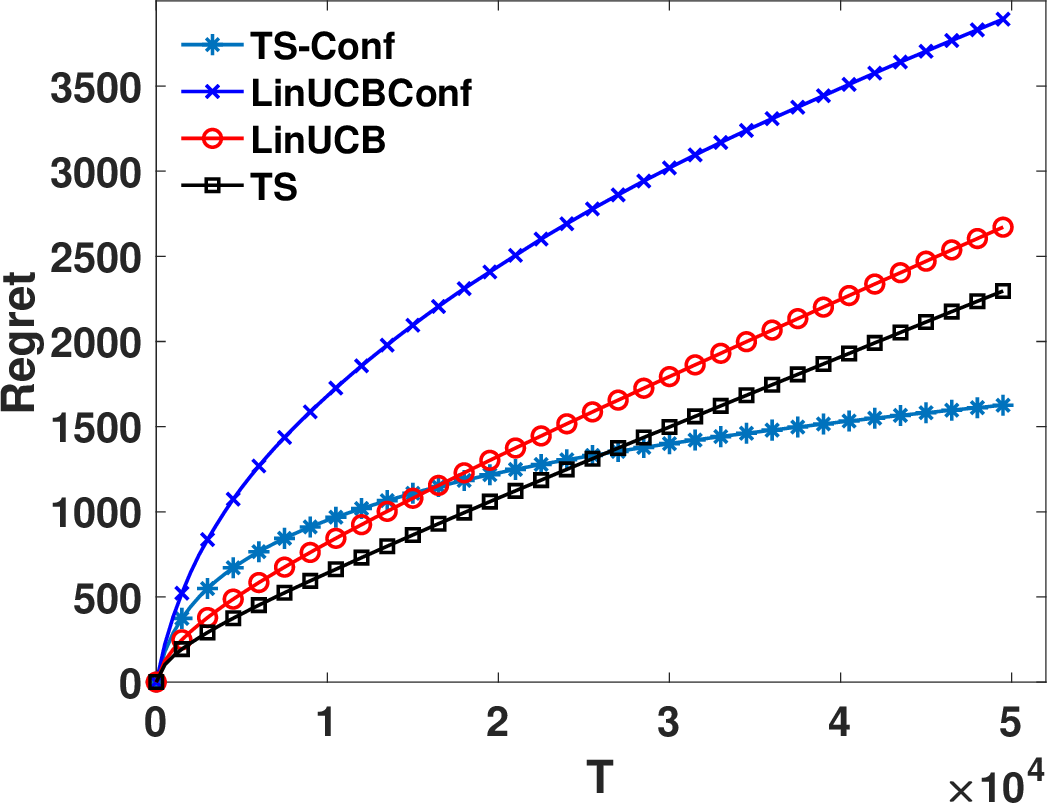}
\subcaption{Dimension $d=20$}
\end{subfigure}
\centering
\begin{subfigure}[b]{0.24\textwidth}
\includegraphics[width=\textwidth]{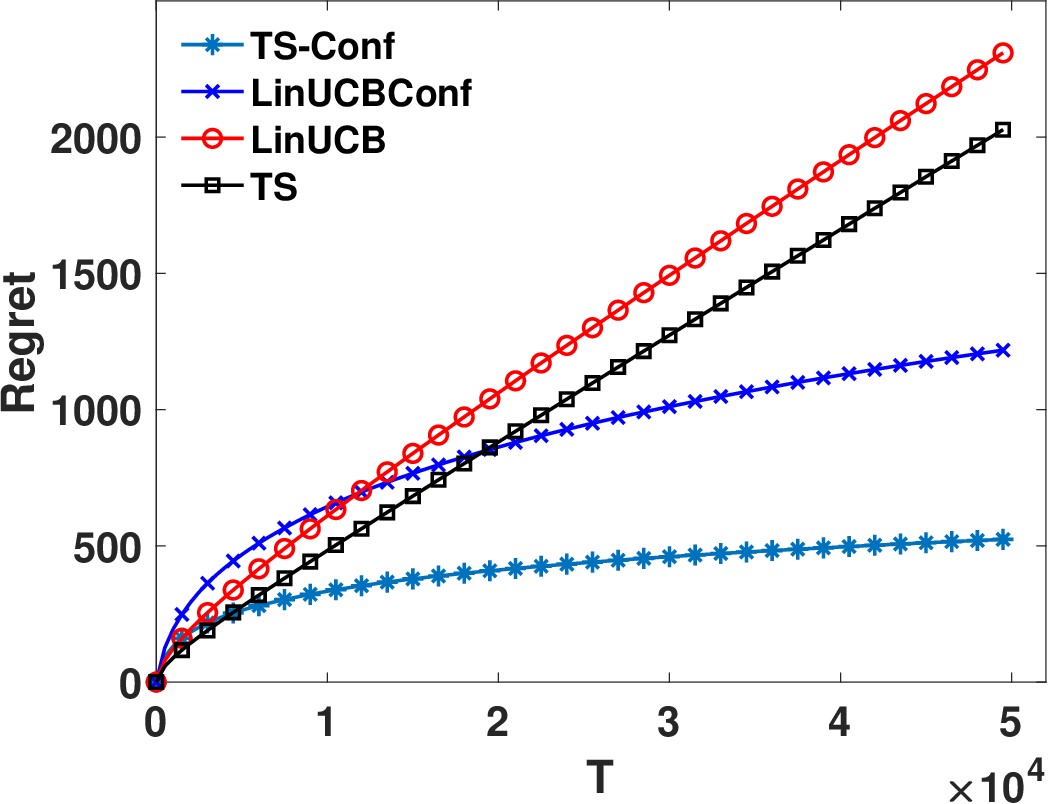}
\subcaption{Noise Var $\sigma^2=0.5$}
\end{subfigure}
\begin{subfigure}[b]{0.24\textwidth}
\includegraphics[width=\textwidth]{pic/Alg1_noise_in/Yelp/algo_contrast/regret/dimension_10_noise_1.0.eps}
\subcaption{Noise Var $\sigma^2=1.0$}
\end{subfigure}
\begin{subfigure}[b]{0.24\textwidth}
\includegraphics[width=\textwidth]{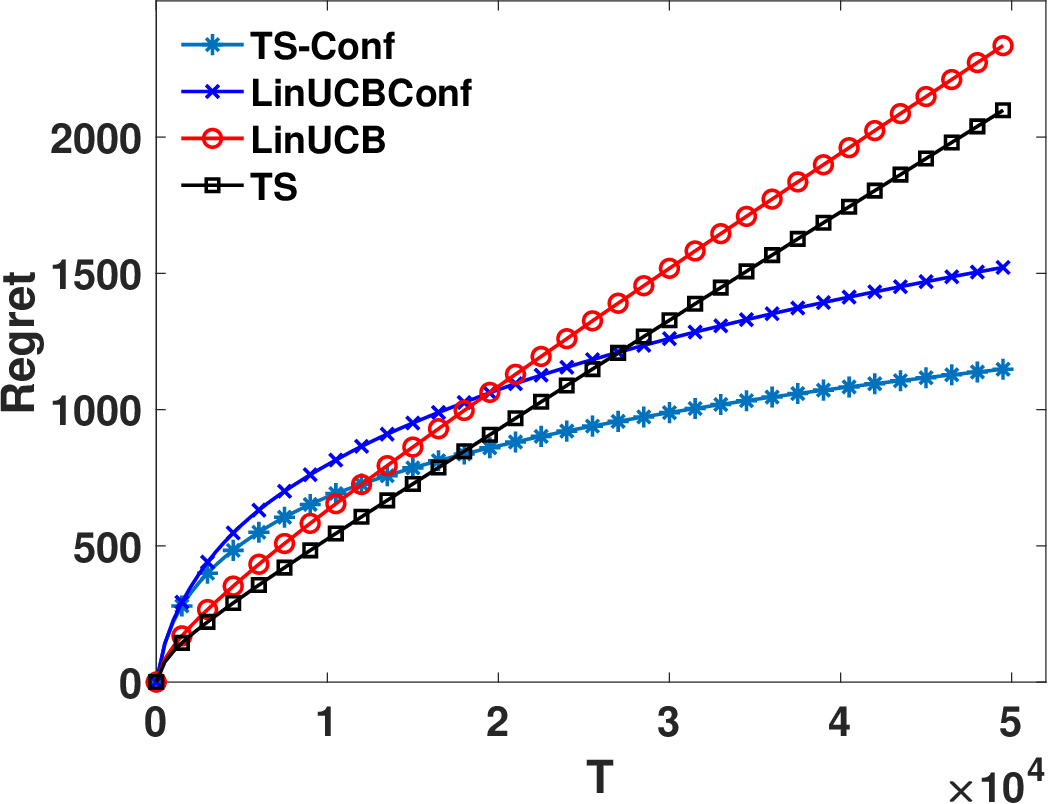}
\subcaption{Noise Var $\sigma^2=1.5$}
\end{subfigure}
\begin{subfigure}[b]{0.24\textwidth}
\includegraphics[width=0.99\textwidth]{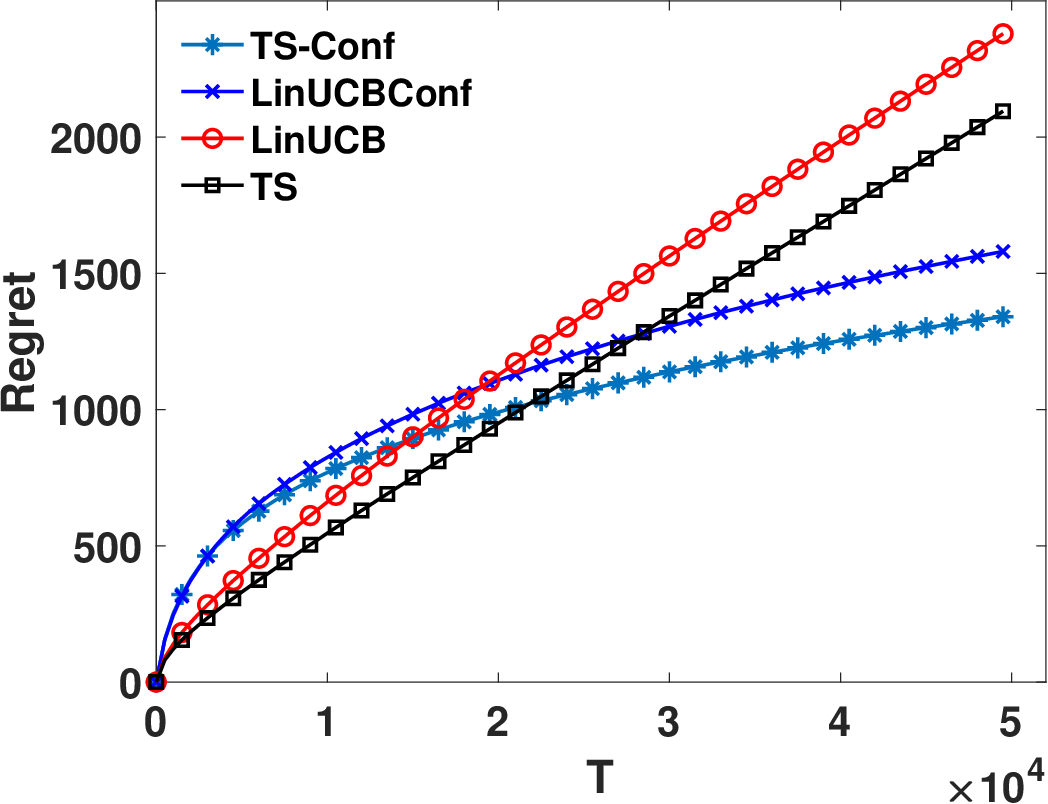}
\subcaption{Noise Var $\sigma^2=2.0$}
\end{subfigure}
\caption{Impact of dimensions $d$ and noise variance $\sigma^2$ in Yelp dataset.
} 
\label{fig16}
\end{figure}

\noindent \textbf{Results in Yelp.} 
Figure \ref{fig16} shows the comparison results of 
the four algorithms on the Yelp dataset. 
It is evident that the TS-Conf algorithm consistently 
has the lowest regret value across different dimensions 
and noise levels. Unlike the LinUCB and TS algorithms, 
which consistently exhibit linear regret growth, 
the regret of TS-Conf gradually converges over time. 
Furthermore, its convergence speed is greater than that of LinUCBConf. \textit{Similar results for Amazon Music and Google Maps datasets are presented in our technical report\cite{xu2024}.} 

\section{Conclusion}
This paper presents a contextual bandit framework to 
address the herding effects problem in recommendation 
applications. In this framework, we assume that the 
user feedback on the action is biased and influenced 
by user preferences and the historical ratings of 
this action. We design a TS-Conf algorithm that 
leverages a posterior sampling technique to 
effectively balance the trade-off between exploration 
and exploitation. 
Our theoretical analysis established a sublinear regret bound for the TS-Conf algorithm, demonstrating its efficiency. 
Extensive experiments on synthetic and real-world datasets show that TS-Conf consistently outperforms three benchmark algorithms, confirming its robustness and superior performance in handling herding effect-induced biases.

\bibliographystyle{splncs04}
\bibliography{reference_st}

\section*{Appendix}

\noindent
{\bf Proof of Theorem \ref{RegretUpper}: } 
Given an arm $a$, one can have: 
\begin{align*}
& (1-\alpha)| \bm{x}^T \bm{\theta} - \bm{x}^T_a \widehat{\bm{\theta}}_t | \\
& = \left| \begin{array}{c c} [\bm{x}^T_a & 0] \end{array} \left[ \begin{array}{c} (1-\alpha)(\bm{\theta} - \widehat{\bm{\theta}}_t) \\ 0 \end{array} \right] \right| \\
& = \left| \begin{array}{c c} [\bm{x}^T_a & 0] \end{array} \left[ \begin{array}{c} (1-\alpha) \bm{\theta} - (1-\alpha) \widehat{\bm{\theta}}_t \\ \alpha - \widehat{\alpha}_t \end{array} \right] \right| \\
& = \left| ( \begin{array}{c c} [\bm{x}^T_a & h_{t,a}] \end{array} - \begin{array}{c c} [\bm{0} & h_{t,a}] \end{array} ) \left[ \begin{array}{c} (1-\alpha) \bm{\theta} - (1-\alpha) \widehat{\bm{\theta}}_t \\ \alpha - \widehat{\alpha}_t \end{array} \right] \right| \\
& = \left| \begin{array}{c c} [\bm{x}^T_a & h_{t,a}] \end{array} \left[ \begin{array}{c} (1-\alpha) \bm{\theta} - (1-\alpha) \widehat{\bm{\theta}}_t \\ \alpha - \widehat{\alpha}_t \end{array} \right] \right| - \left| \begin{array}{c c} [\bm{0} & h_{t,a}] \end{array} \left[ \begin{array}{c} (1-\alpha)\bm{\theta} - (1-\alpha)\widehat{\bm{\theta}}_t \\ \alpha - \widehat{\alpha}_t \end{array} \right] \right| \\
& \leq \left| \begin{array}{c c} [\bm{x}^T_a & h_{t,a}] \end{array} \left[ \begin{array}{c} (1-\alpha)\bm{\theta} - (1-\alpha)\widehat{\bm{\theta}}_t \\ \alpha - \widehat{\alpha}_t \end{array} \right] \right| + \left| \begin{array}{c c} [\bm{0} & h_{t,a}] \end{array} \left[ \begin{array}{c} (1-\alpha)\bm{\theta} - (1-\alpha)\widehat{\bm{\theta}}_t \\ \alpha - \widehat{\alpha}_t \end{array} \right] \right| \\
& \leq \left| \begin{array}{c c} [\bm{x}^T_a & h_{t,a}] \end{array} \left[ \begin{array}{c} (1-\alpha)\bm{\theta} - (1-\alpha)\widehat{\bm{\theta}}_t \\ \alpha - \widehat{\alpha}_t \end{array} \right] \right| + h_{t,a} | \alpha - \widehat{\alpha}_t |.
\end{align*}

Note that the first term corresponds to the linear regression problem 
of estimating the parameter $[(1-\alpha) \bm{\theta}^T \alpha]^T$, 
with observation: 
\begin{equation}
	\begin{aligned}
		V_{t}(a)
		&=\left[h_{t,a};
		\boldsymbol{x}^T_{a}
		\right] 
		\left[\begin{array}{c}
			\alpha \\ 
			(1-\alpha)\boldsymbol{\theta}
		\end{array}\right]+\eta_{t}.
	\end{aligned}
	\label{equa8}
	\nonumber
\end{equation}
The second term corresponding to estimating the strength of herding effects $\alpha$.  
Thus, the upper confidence bound of $(1-\alpha)|
\bm{x}^T \bm{\theta} 
- 
\bm{x}^T_a \widehat{\bm{\theta}}_t
|$  
can be bounded by the upper confidence bound of the linear regression problem of estimating 
$[(1-\alpha) \bm{\theta}^T \alpha]^T$ plus the upper confidence bound of estimating $\alpha$.  
Apply \cite{2014russo} and \cite{Lattimore2020}, the Bayesian regret can be bounded as 
\begin{align*}
R^{Bay}_T(\mathcal{D})  
&\leq 
O\left( 
\frac{1}{1-\alpha} 
\int 
\sum^T_{t=1} W^\ast_t (1/T; \bm{\Psi}) d \bm{\Psi} \right. + \left. \frac{1}{1-\alpha} d \sqrt{T} \ln T 
\right).
\end{align*}
This proof is then complete.  
\done

\noindent
{\bf Proof of Theorem \ref{RegetLower}: }
Note that $\alpha \in [0,1]$, implying that $W^\ast_t (1/T; \bm{\Psi}) \leq 1$.  
The case $\sum^T_{t=1} W^\ast_t (1/T; \bm{\Psi})  = \Omega(T)$ implies 
that there exists a constant $c$ such that $|\alpha - \widehat{\alpha}_t| \geq c$ 
hold for $ \Omega(T)$ rounds for any $\bm{\Psi}$.  
One can select instances of $\bm{\Psi}$ such that the gap between the optimal 
arm and the sub-optimal arm with the largest the reward is larger or equal to $c$.  
In such instances, the optimal arm the sub-optimal arm is indistinguishable.  
Resulting a linear regret of $ \Omega(T)$.  
Defining the prior distribution over such instances, leads to a Bayesian regret 
of $ \Omega(T)$. 
\done

\noindent 
{\bf A special case.}  
We consider an important special case 
where exact sampling from the posterior can be computationally efficient.  
Equation (\ref{equq_reward})  
can be rewritten as follows: 
$
V_{t}(A_t)
=
\alpha h_{t,{A_t}}+(1-\alpha)
\boldsymbol{\theta}^{\mathrm{T}} 
\boldsymbol{x}_{A(t)}+\eta_{t}
=\widetilde{\boldsymbol{x}}_{A(t)}^{\text{T}} 
\widetilde{\boldsymbol{\theta}}+\eta_{t, A_t}
$
where 
$\widetilde{\boldsymbol{\theta}} =
\left[\begin{array}{c}\alpha \\ 
(1-\alpha)\boldsymbol{\theta}
\end{array}\right]$ and 
$\widetilde{\boldsymbol{x}}_{A(t)}=
\left[h_{t,{A_t}};
\boldsymbol{x}_{A(t)}
\right]$. 
The special case is composed of Gaussian distributions:  
(1) the priors of 
the $\widetilde{\boldsymbol{\theta}} $ 
follows multivariate Gaussian 
distribution with 
mean $\boldsymbol{\mu}$ and 
covariance $\boldsymbol{\Lambda^{-1}}$, 
i.e. $p(\widetilde{\boldsymbol{\theta}} )
\sim \mathcal{N}(\boldsymbol{\mu},
\boldsymbol{\Lambda^{-1}})$; 
(2) the noise $\eta_{t}$ follows 
a Gaussian distribution, 
i.e., $f(\eta, \sigma_a)$ is 
the density function of 
$\mathcal{N}(0, \sigma^2_a)$; 
(3) the variance $\sigma^2_a$ is given. 
Under this special case, 
the posterior $p\left(\widetilde{\boldsymbol{\theta}} |
\mathcal{H}_{t}\right)$ 
follows $\mathcal{N}(\boldsymbol{\mu_t}, \boldsymbol{\Sigma_t})$, where 
$
\boldsymbol{\Sigma_{t}}
=\Bigl(
\boldsymbol{\Lambda}
+\frac{1}{\sigma_{n}^{2}}
{\textstyle \sum_{\tau=1}^{t-1}}
\widetilde{\boldsymbol{x}}_{A(\tau)}
\cdot(\widetilde{\boldsymbol{x}}
_{A(\tau)})^{\mathrm{T}}
\Bigr)^{-1},
$
$
\boldsymbol{\mu_{t}}
=\boldsymbol{\Sigma_{t}}
(\frac{1}{\sigma_{n}^{2}}
{\textstyle \sum_{\tau=1}^{t-1}}
V_{\tau}{(A_\tau)}
\cdot 
\widetilde{\boldsymbol{x}}_{A(\tau)}
+\boldsymbol{\Lambda}
\boldsymbol{\mu}).  
$  
These formulas imply that sampling the posterior  $p\left(\widetilde{\boldsymbol{\theta}} |
\mathcal{H}_{t}\right)$ is sampling from Gaussian distributions. 
A sample of $\bm{\theta}$ can be obtained from a sample of $\widetilde{\boldsymbol{\theta}}$ as 
follows:  
$\bm{\theta} = \widetilde{\boldsymbol{\theta}}_{[2:d+1]} / (1 - \widetilde{\theta}_1)$, 
where $\widetilde{\boldsymbol{\theta}}_{[2:d+1]}$ denotes a vector composed of the 
second to the $(d+1)$-th entries of $ \widetilde{\boldsymbol{\theta}}$ 
and $\widetilde{\theta}_1$ denotes the first entry of $ \widetilde{\boldsymbol{\theta}}$.  

\noindent \textbf{Implementation details.}
For synthetic datasets, we generate the observed action features 
and the user preference vector from  
Gaussian distributions, represented as 
$\mathcal{N}(\boldsymbol{\mu}, \boldsymbol{\Sigma})$. 
Here, $\boldsymbol{\mu}$ 
is a $d$-dimensional mean vector with 
a value of $\frac{1}{2}$, 
and $\boldsymbol{\Sigma}$ is a 
$d{\times}d$ covariance matrix with 
diagonal elements set to $\frac{1}{6}$. 
And, the user conformity tendency, 
denoted as $\alpha$, and the historical rating 
of action, represented as $h_{t,a}$, 
are assumed to lie within the intervals $[0,1]$ 
and $[0,5]$, respectively. 
This approach ensures that our simulations 
closely resemble real-world scenarios, 
making them both believable and directly 
applicable to actual data patterns. 

For real-world datasets, the Amazon Music dataset encompasses 
user ratings for musical items available on Amazon; 
The MovieLens dataset captures user ratings for 
movies hosted on the MovieLens platform; 
The Yelp dataset aggregates user ratings 
for restaurants listed on Yelp; 
The Google Maps dataset collates ratings 
and reviews for various locations and 
venues available on Google Maps. Table~\ref{tb:dataset_stats} presents the 
detailed statistics of these datasets. 

\begin{table}[htb]
\caption{Datasets Summary} 
\centering
\begin{tabular}{cccc}
\hline
\label{tb:dataset_stats}
\textbf{Datasets} & \textbf{Users} & \textbf{Items} & \textbf{Ratings} \\ \hline
MovieLens        & 6,040          & 3,706          & 1,000,209        \\
Amazon music     & 478,235        & 266,414        & 836,005          \\
Google map       & 5,054,567      & 3,116,785      & 11,453,845       \\
Yelp             & 366,715        & 60,785         & 1,569,264        \\ \hline
\end{tabular}
\end{table}

We conduct our experiments with 
$T = 50000$ decision rounds. 
In each round, all actions($|\mathcal{A}| = 10$) 
are presented to the decision maker, 
denoted as $\mathcal{A}_t = \mathcal{A}$ 
for all $t \in [T]$. 
The noise in the agent's feedback is 
simulated according to Eq. (2) 
using a normal distribution with 
variance $\sigma^2_a=1.0$. 
The density function $f(\eta, \sigma_a=1.0)$ 
corresponds to $\mathcal{N}(\eta, 1)$. 
By default, the feature dimensions are set to 
$d = 10$, unless otherwise specified. 
The prior distribution of user preference vector follow  
Gaussian distributions, represented as 
$\mathcal{N}(\boldsymbol{\mu}, \boldsymbol{\Sigma})$. 
Here, $\boldsymbol{\mu}$ 
is a $d$-dimensional mean vector with 
a value of $0$, 
and $\boldsymbol{\Sigma}$ is a 
$d{\times}d$ covariance matrix with 
diagonal elements set to $1$. 
The $h_{t,a}$ is set as the average rating of an item.  
Lastly, prior distribution of $\alpha$ follows $\mathcal{N}(0,1)$. 

\noindent \textbf{Data Processing Details of Real-World Datasets.}
To ensure the integrity of the experiment 
and mitigate the impact of missing values, 
we apply certain data filtering criteria. 
Specifically, we exclude items with a rating count of
less than 10 and users with a rating count of 
less than 10. This filtering process allows 
us to focus on constructing the necessary dataset 
for our analysis. To begin with, we calculate the 
average score for each action $a$ in the dataset, 
which serves as the historical score $r^*_a$ 
for that action. Additionally, we employ 
the Matrix Factorization (MF) technique 
to learn item features $\boldsymbol{x}_a$, 
user features $\boldsymbol{\theta}$, 
and user conformity tendency $\alpha$. 
In the case of herding effects, 
the MF rating model is:
$
\hat{r}_{u i} = 
(1-\texttt{Sig}(\beta)) 
\boldsymbol{\theta}_{u}^{T} 
\cdot \boldsymbol{x}_{i}+
\texttt{Sig}(\beta)
r^*_a, 
$
where $\texttt{Sig}()$ is the sigmoid function, 
$\texttt{Sig}(\beta)$ represents the 
estimated value of user 
conformity tendency $\alpha$. 
In the model MF, for each user $u$ and item $i$, 
by comparing the predicted score $\hat{r}^*_{ui}$ 
with the real score $r^*_{ui}$ difference 
to update parameters. We learn 
the variables $\alpha$, $\boldsymbol{x}$ and 
$\boldsymbol{\theta}$ from five dimensions: 
$d = 5$, $d = 10$, 
$d = 15$, $d = 20$ respectively.   
We input these inferred variables to the reward model, i.e., Eq. (2) to generate the reward.  
The noise in the reward follows a normal distribution with 
variance $\sigma^2_a=1.0$.   

\begin{figure}
\centering
\begin{subfigure}[b]{0.3\textwidth}
\includegraphics[width=\textwidth]{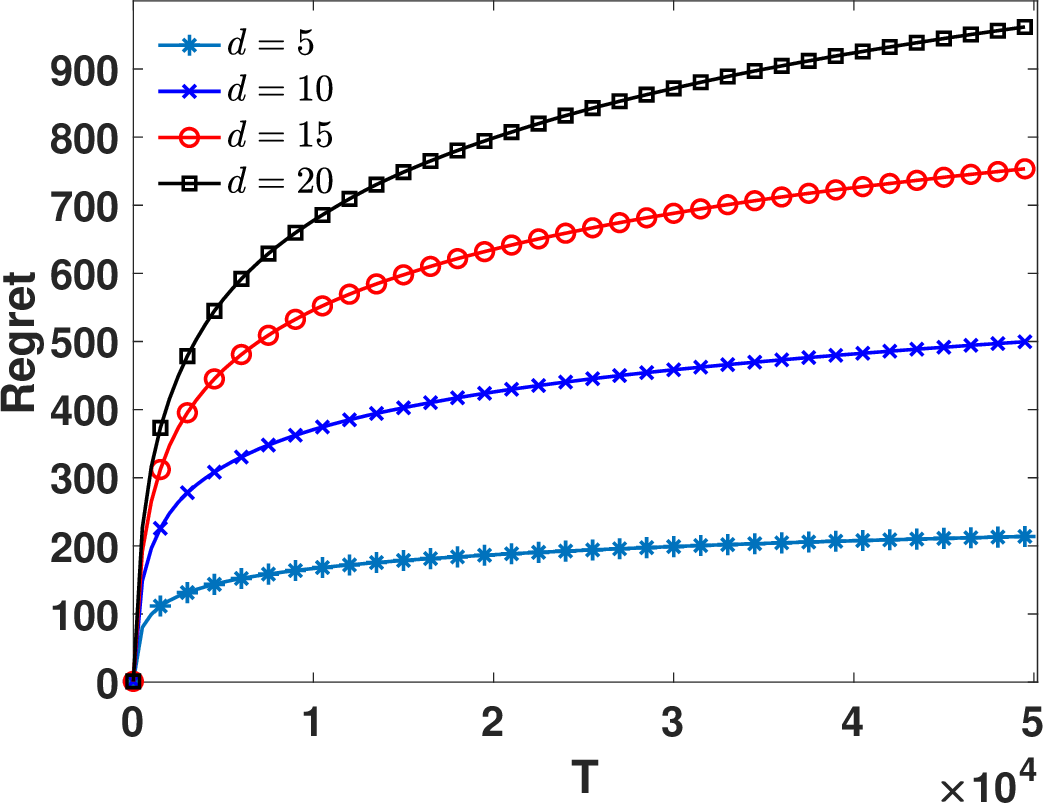}
\subcaption{Impact of dimensions $d$}
\label{fig3(a)}
\end{subfigure}
\begin{subfigure}[b]{0.3\textwidth}
\includegraphics[width=\textwidth]{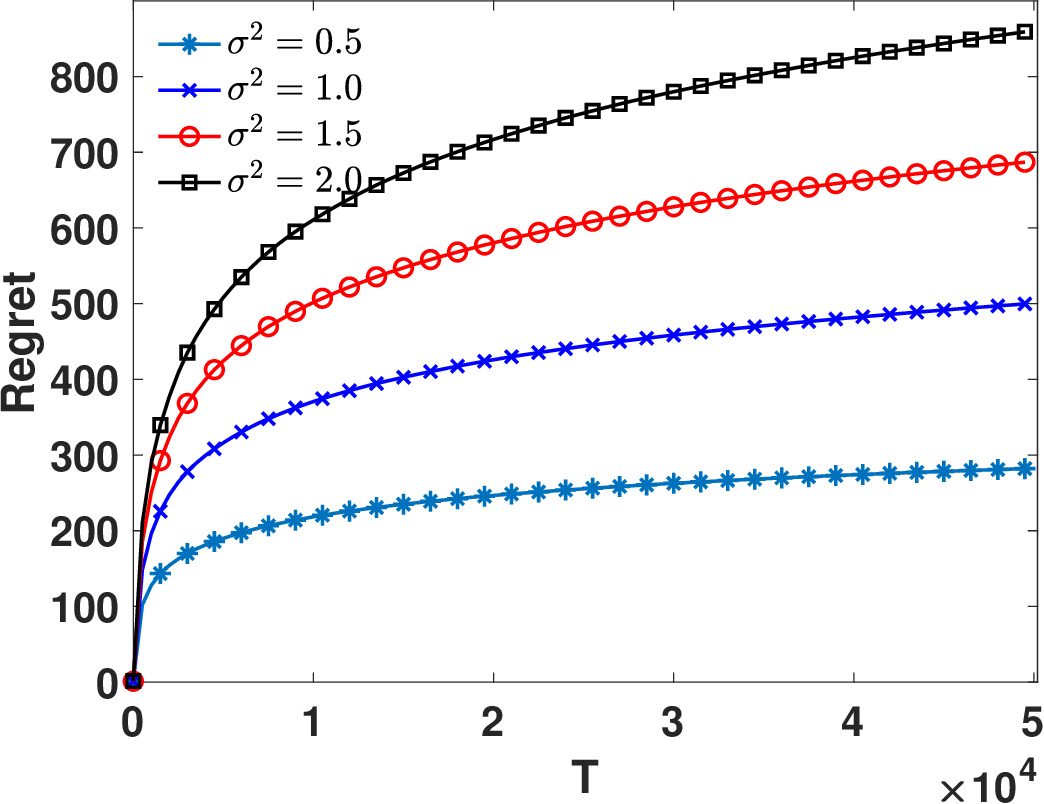}
\subcaption{Impact of noise var $\sigma^2$}
\label{fig3(b)}
\end{subfigure}
\caption{TS-Conf's performance on different 
$d$ and $\sigma^2$
} 
\label{fig3}
\end{figure}

\noindent \textbf{Convergence Analysis.} 
This experiment examines the convergence behavior 
of the proposed TS-Conf algorithm under 
varying parameters, notably dimensionality $d$ 
and noise variance $\sigma^2$. 
Figure \ref{fig3(a)} benchmarks TS-Conf's 
convergence across distinct feature dimensions, 
with $d$ values set to [5, 10, 15, 20], all the 
while maintaining a consistent noise variance of 
$\sigma^2 = 1.0$. Conversely, Figure \ref{fig3(b)} 
evaluates its convergence under a spectrum of noise 
variances, specifically 
$\sigma^2 = [0.5, 1.0, 1.5, 2.0]$, 
with the dimensionality held constant at $d= 10$.
A clear pattern emerges: 
TS-Conf consistently 
posts the lowest regret 
values at the minimal settings of $d$ and 
$\sigma^2$. As these parameters escalate, the regret 
correspondingly surges. This behavior suggests that 
the algorithm grapples more with the 
exploration-exploitation trade-off as 
either $d$ or $\sigma^2$ amplifies. 
This observation aligns our derived regret 
bound presented in Theorem \ref{RegretUpper}. 
Notably, TS-Conf excels in scenarios with 
partially observable features or diminished uncertainty levels. 

\begin{figure}
\centering
\begin{subfigure}[b]{0.24\textwidth}
\includegraphics[width=\textwidth]{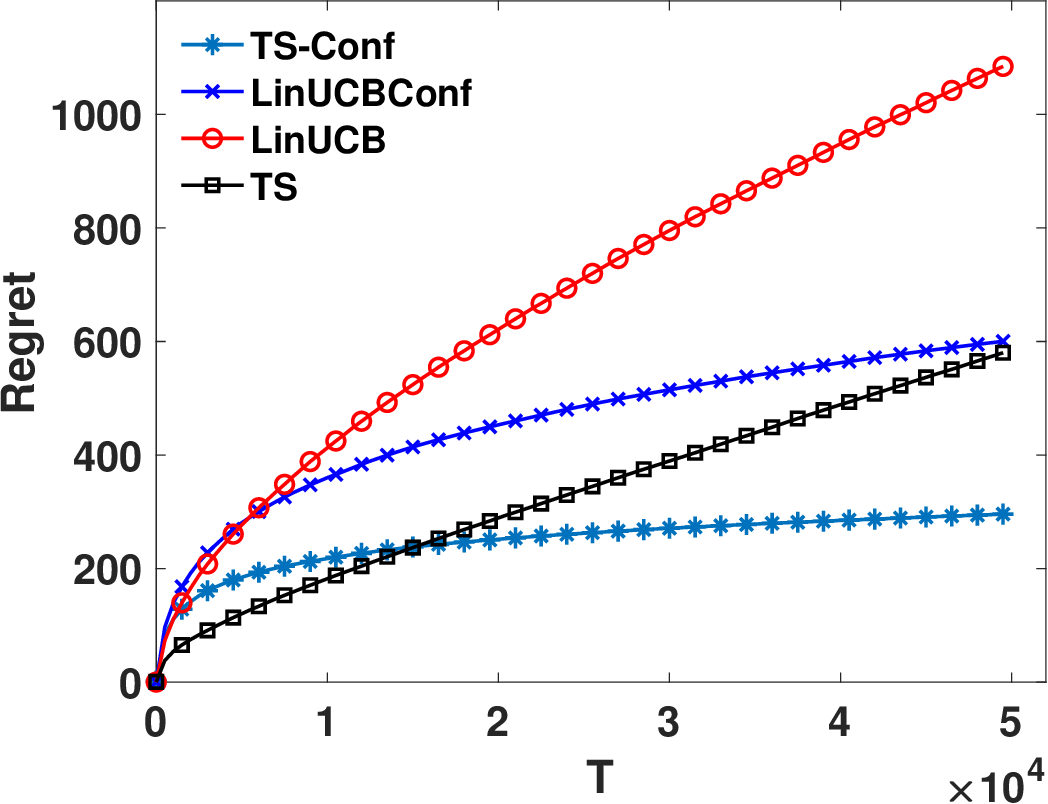}
\subcaption{Dimension $d=5$}
\end{subfigure}
\begin{subfigure}[b]{0.24\textwidth}
\includegraphics[width=\textwidth]{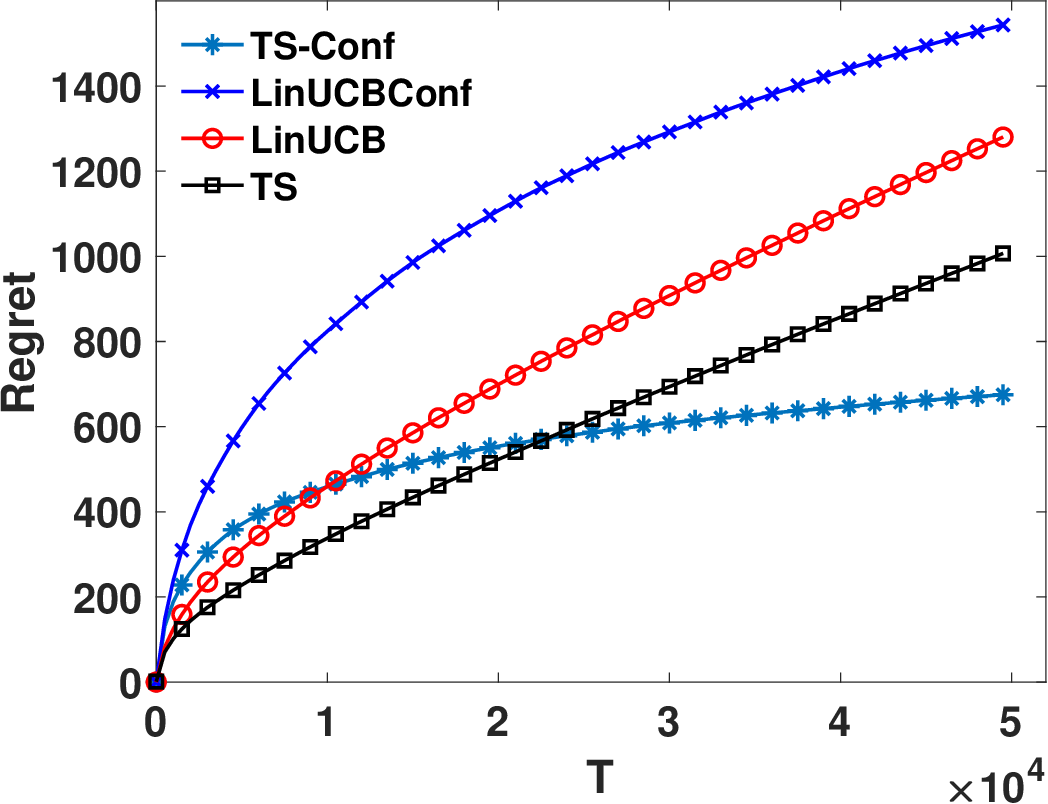}
\subcaption{Dimension $d=10$}
\end{subfigure}
\begin{subfigure}[b]{0.24\textwidth}
\includegraphics[width=\textwidth]{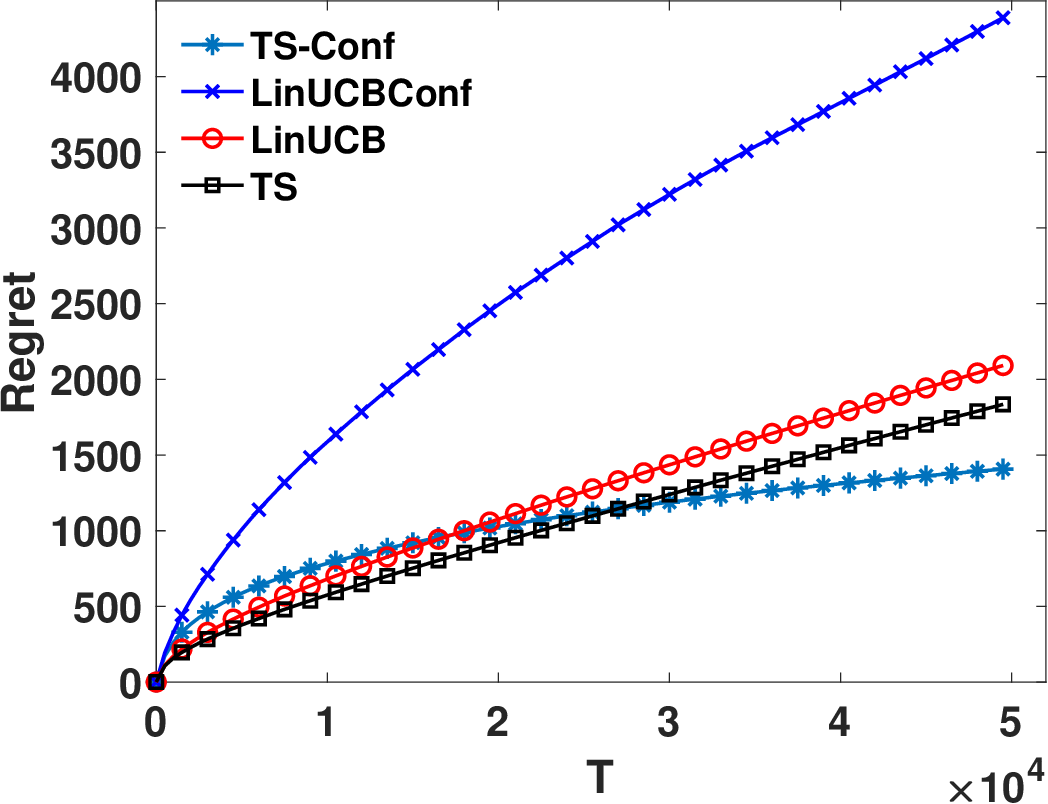}
\subcaption{Dimension $d=15$}
\end{subfigure}
\begin{subfigure}[b]{0.24\textwidth}
\includegraphics[width=0.99\textwidth]{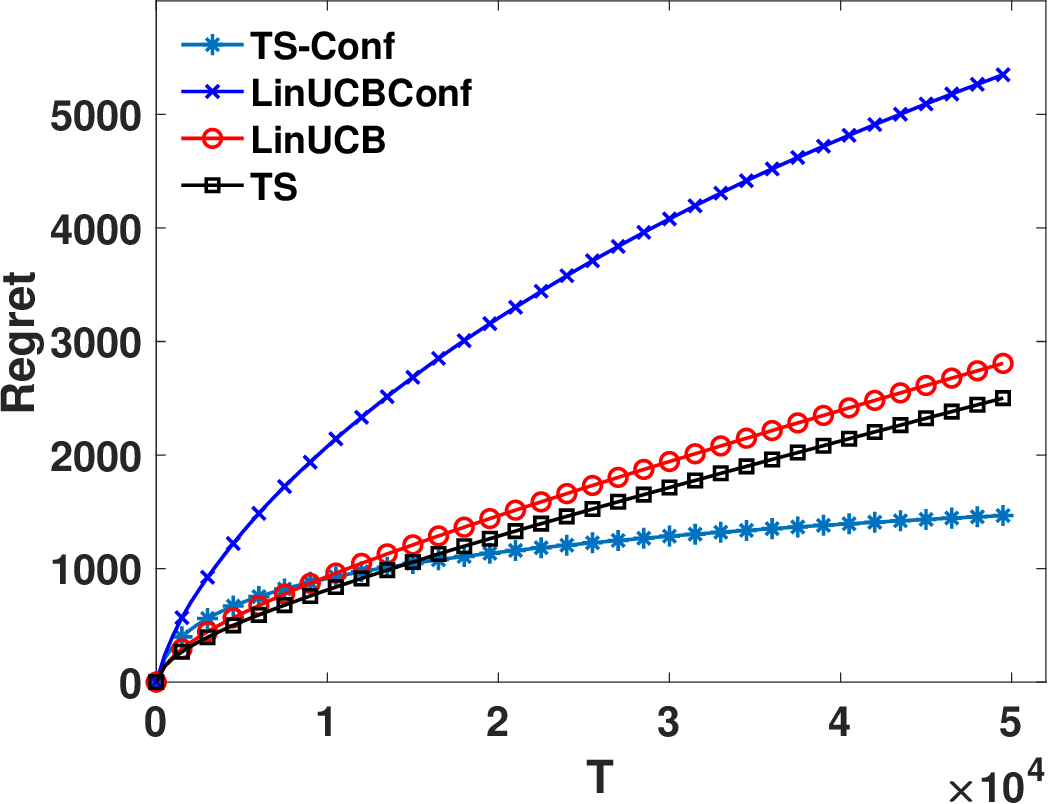}
\subcaption{Dimension $d=20$}
\end{subfigure}
\centering
\begin{subfigure}[b]{0.24\textwidth}
\includegraphics[width=\textwidth]{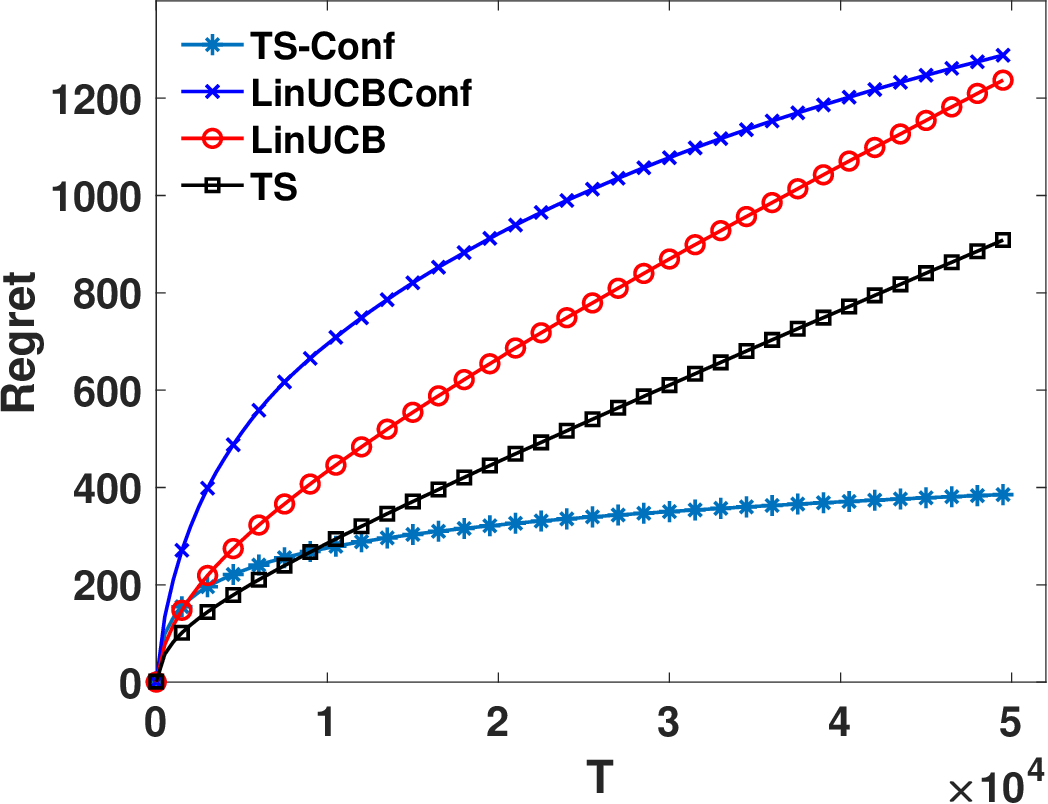}
\subcaption{Noise Var $\sigma^2=0.5$}
\end{subfigure}
\begin{subfigure}[b]{0.24\textwidth}
\includegraphics[width=\textwidth]{pic/Alg1_noise_in/Google_map/algo_contrast/regret/dimension_10_noise_1.0.eps}
\subcaption{Noise Var $\sigma^2=1.0$}
\end{subfigure}
\begin{subfigure}[b]{0.24\textwidth}
\includegraphics[width=\textwidth]{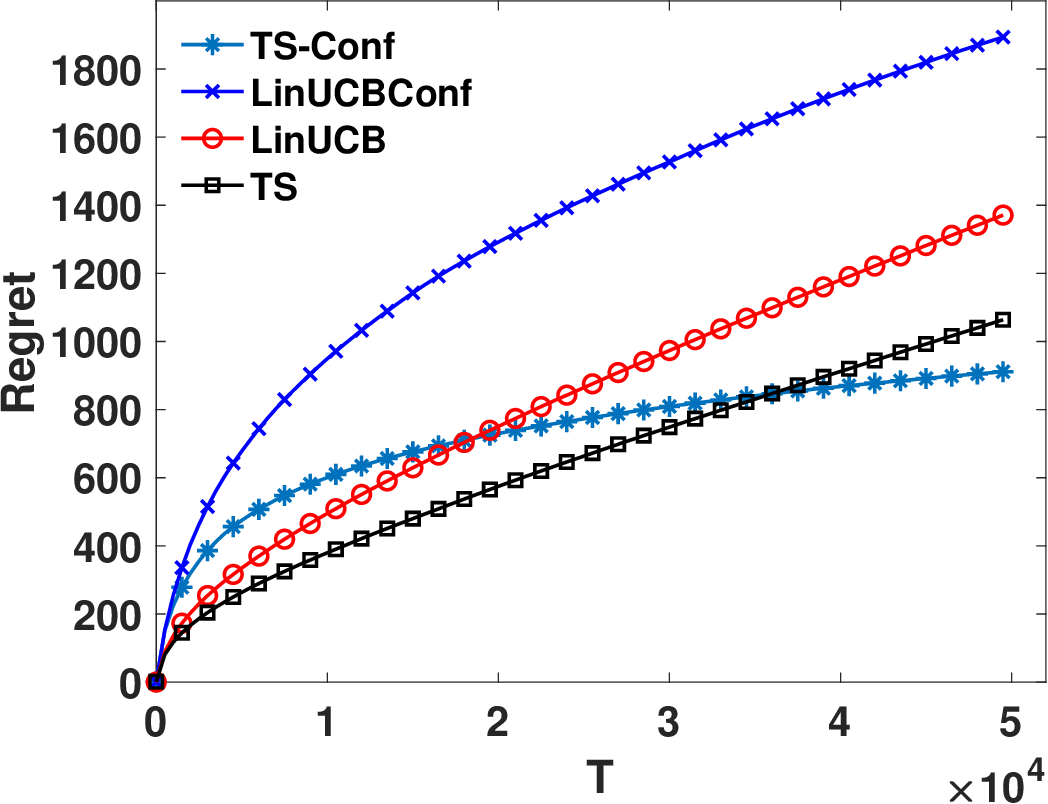}
\subcaption{Noise Var $\sigma^2=1.5$}
\end{subfigure}
\begin{subfigure}[b]{0.24\textwidth}
\includegraphics[width=0.99\textwidth]{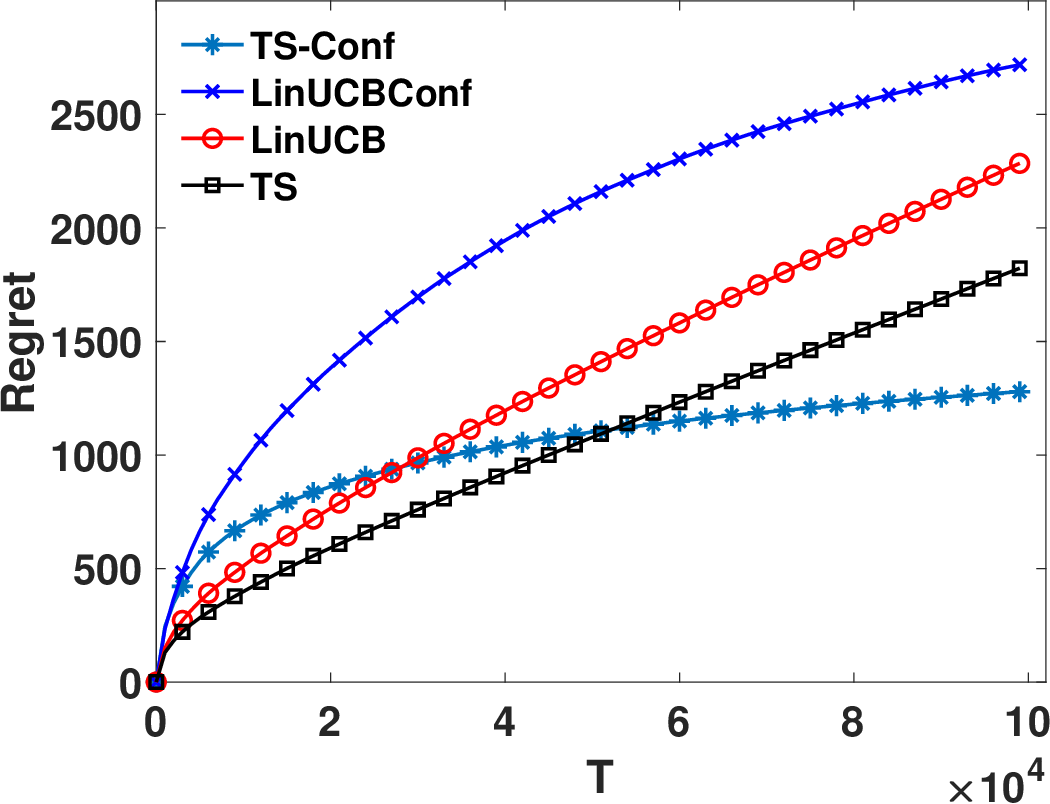}
\subcaption{Noise Var $\sigma^2=2.0$}
\end{subfigure}
\caption{Impact of dimensions $d$ and noise variance $\sigma^2$  
in Google Map dataset.} 
\label{fig19}
\end{figure}

\noindent \textbf{Results in Google Maps.} 
Figure \ref{fig19} shows the comparison results of the 
four algorithms on the Google Maps dataset, respectively. 
Similarly, it is evident that the TS-Conf algorithm always 
has the lowest regret value across different dimensions 
and noise. Differing from the LinUCB 
and TS algorithms, which consistently 
exhibit linear regret growth, the regret of 
TS-Conf gradually converges over time, 
and the convergence speed 
is greater than that of LinUCBConf. 

\begin{figure}
\centering
\begin{subfigure}[b]{0.24\textwidth}
\includegraphics[width=\textwidth]{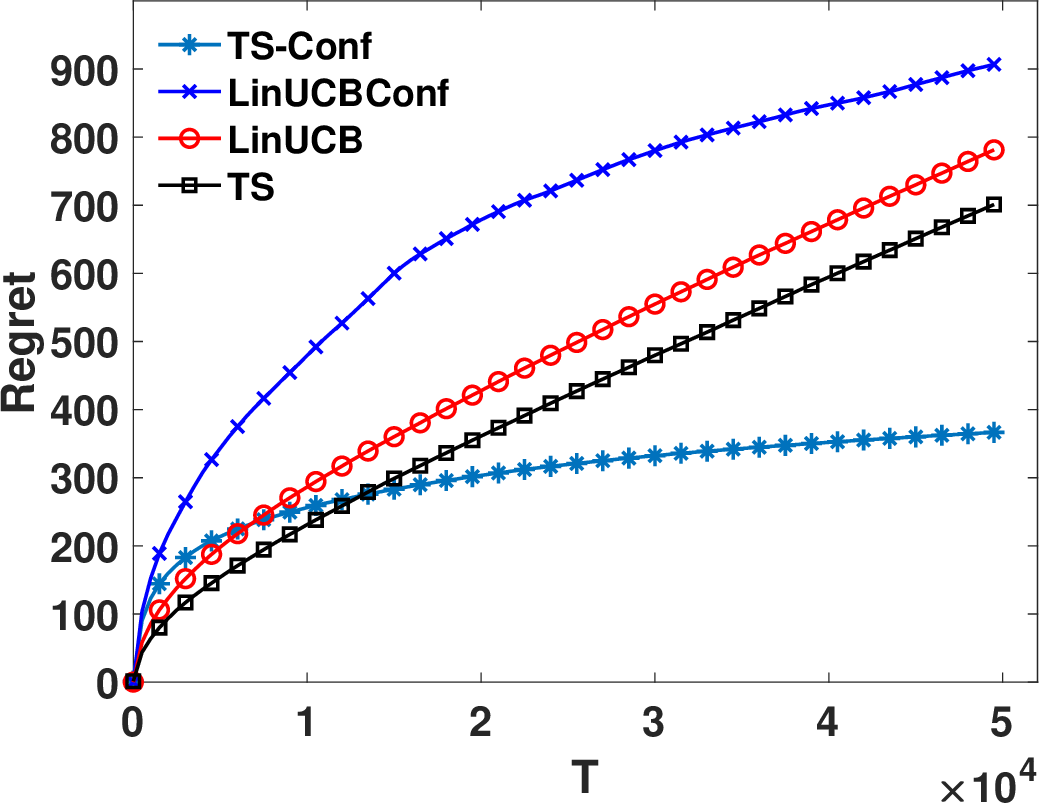}
\subcaption{Dimension $d=5$}
\end{subfigure}
\begin{subfigure}[b]{0.24\textwidth}
\includegraphics[width=\textwidth]{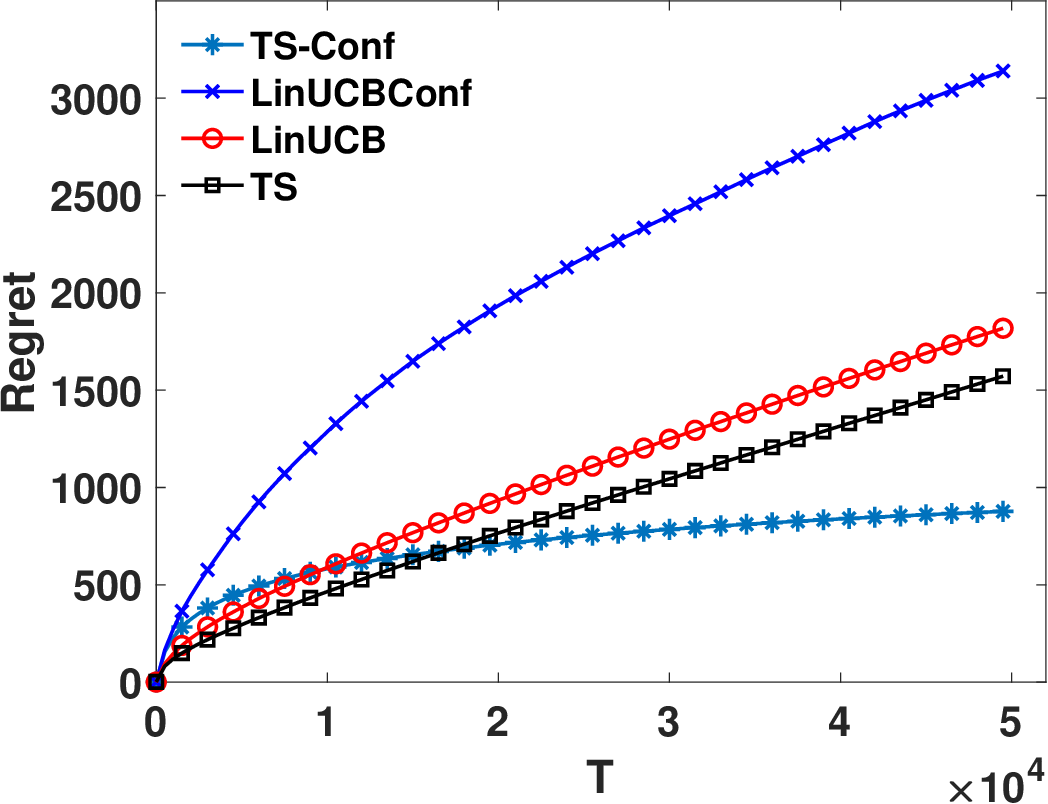}
\subcaption{Dimension $d=10$}
\end{subfigure}
\begin{subfigure}[b]{0.24\textwidth}
\includegraphics[width=\textwidth]{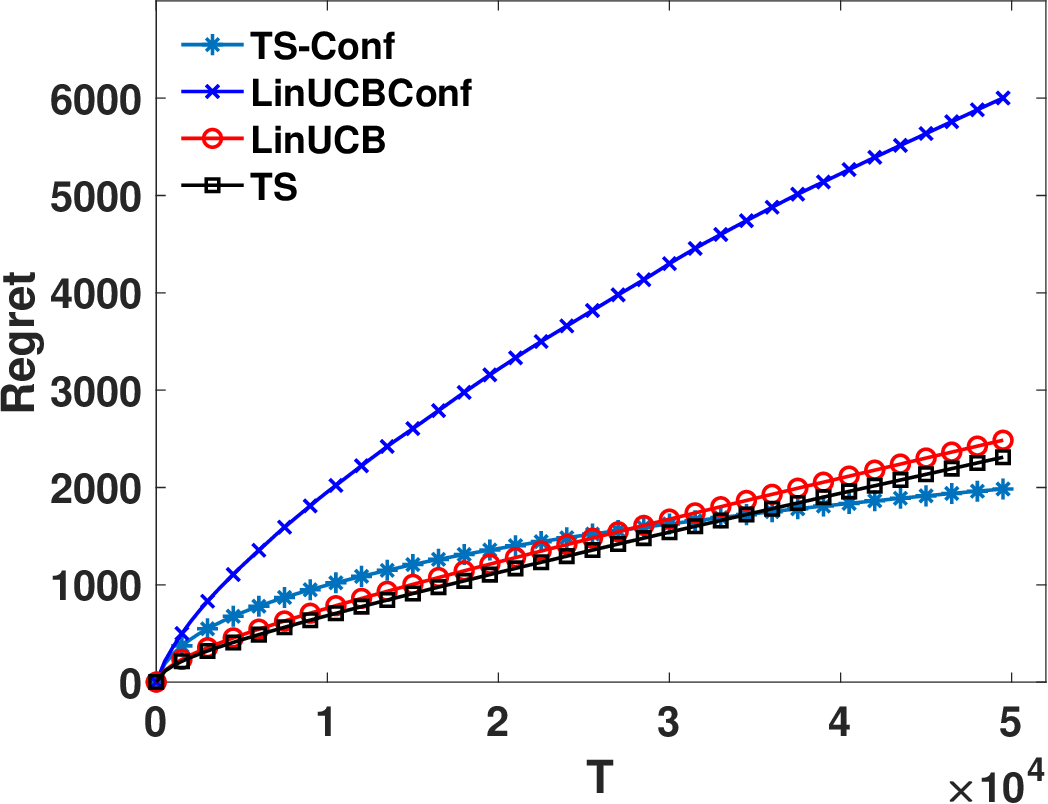}
\subcaption{Dimension $d=15$}
\end{subfigure}
\begin{subfigure}[b]{0.24\textwidth}
\includegraphics[width=0.99\textwidth]{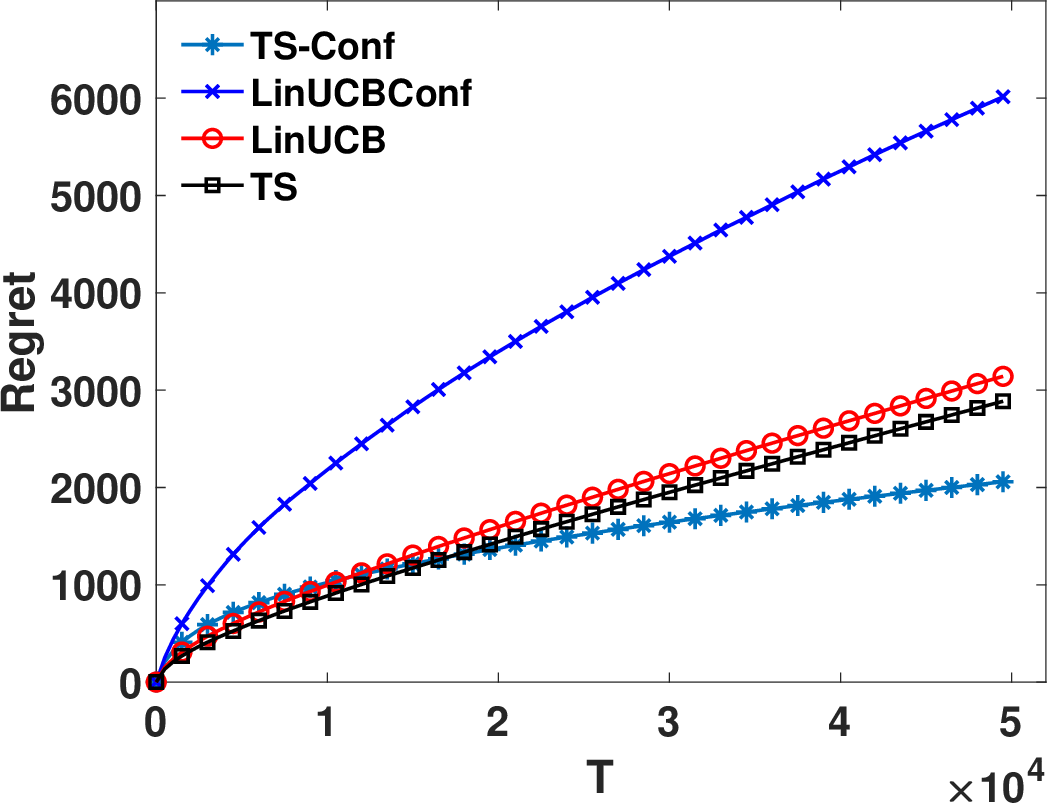}
\subcaption{Dimension $d=20$}
\end{subfigure}
\centering
\begin{subfigure}[b]{0.24\textwidth}
\includegraphics[width=\textwidth]{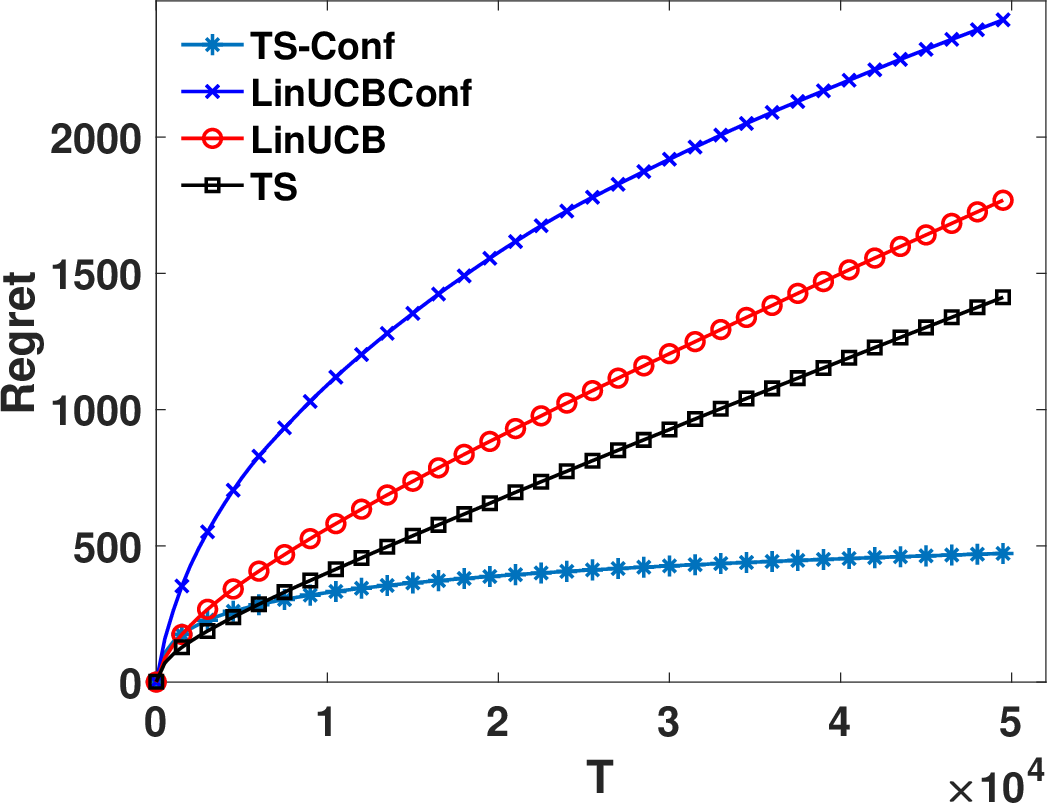}
\subcaption{Noise Var $\sigma^2=0.5$}
\end{subfigure}
\begin{subfigure}[b]{0.24\textwidth}
\includegraphics[width=\textwidth]{pic/Alg1_noise_in/Amazon_music/algo_contrast/regret/dimension_10_noise_1.0.eps}
\subcaption{Noise Var $\sigma^2=1.0$}
\end{subfigure}
\begin{subfigure}[b]{0.24\textwidth}
\includegraphics[width=\textwidth]{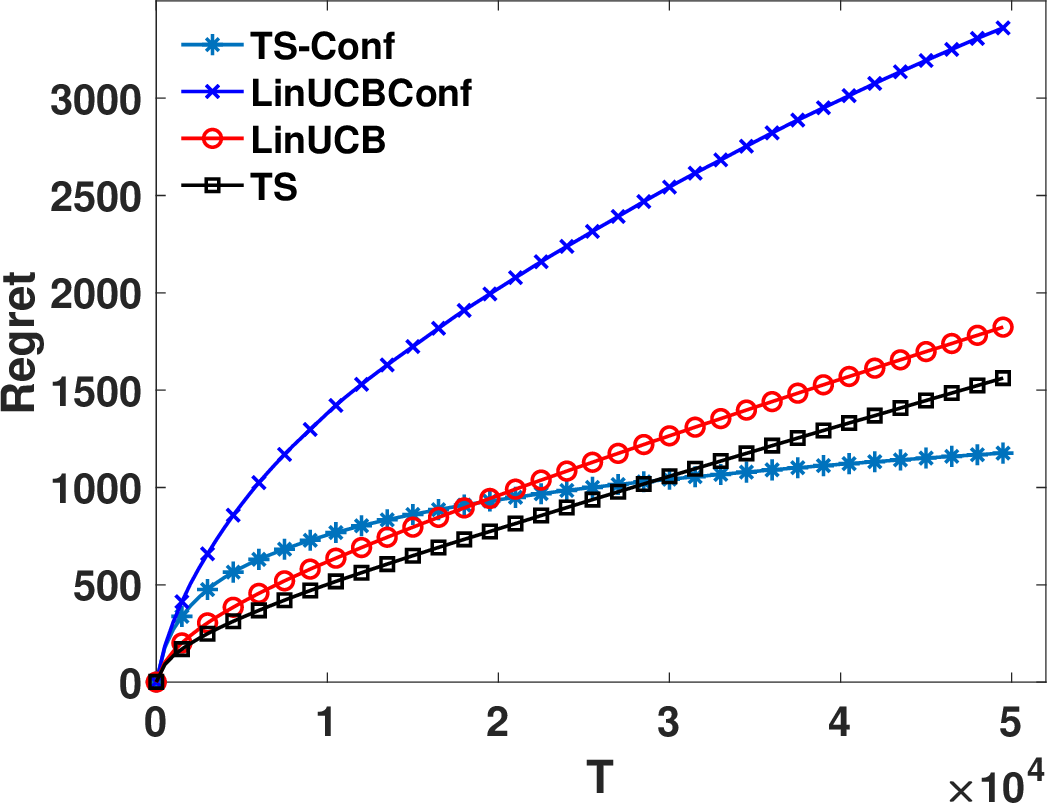}
\subcaption{Noise Var $\sigma^2=1.5$}
\end{subfigure}
\begin{subfigure}[b]{0.24\textwidth}
\includegraphics[width=0.99\textwidth]{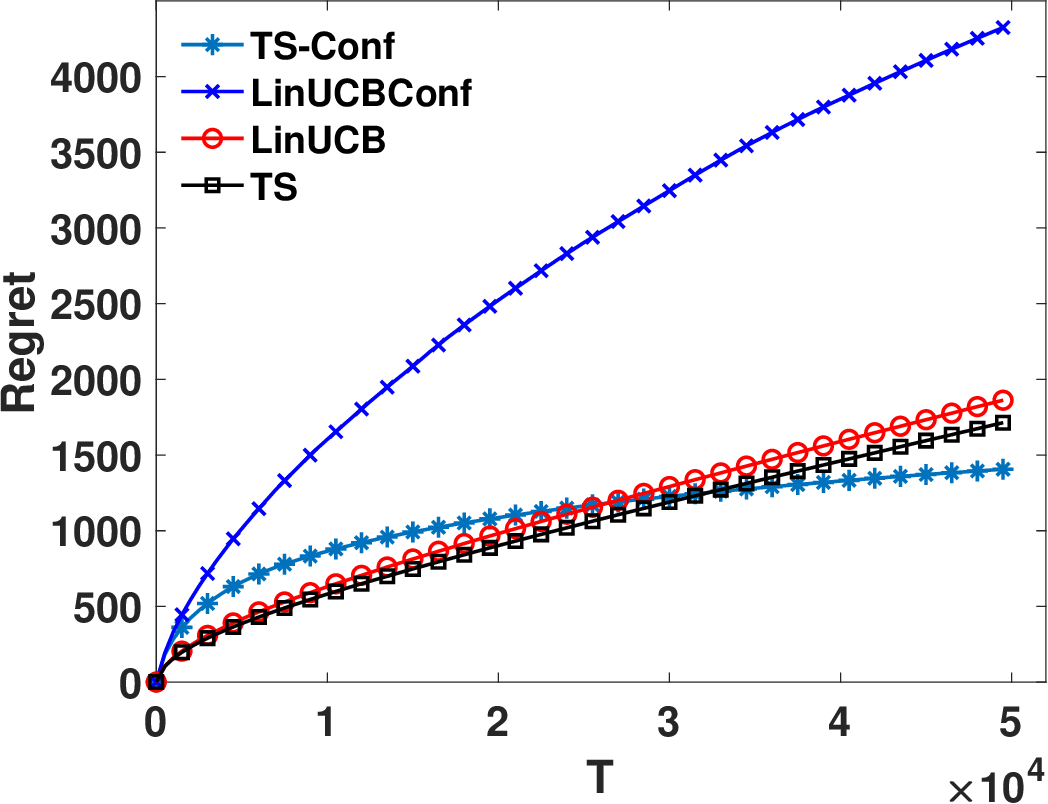}
\subcaption{Noise Var $\sigma^2=2.0$}
\end{subfigure}
\caption{Impact of dimensions $d$ and noise variance $\sigma^2$ 
in Amazon dataset.} 
\label{fig10}
\end{figure}

\noindent \textbf{Results in Amazon Music.}
Figure \ref{fig10} shows the regret $\hat{R}_{t}$
produced by each algorithm in different 
dimensions and different noise variances. 
It can be observed that 
the TS-Conf algorithm 
always has the lowest 
regret value 
across varying dimensions 
and noise levels. In some cases 
(i.e., $d=15, \sigma^2=1.5, \sigma^2=2.0$), 
due to the complexity 
and uncertainty of real scenarios, 
the algorithm's regret 
value in the initial stages 
might be higher than 
the LinUCB and TS algorithms. 
However, it is notable 
that both the LinUCB and TS 
algorithms demonstrate divergence, 
with their regret values 
increasing linearly 
with time ($t$). 
In contrast, the TS-Conf 
algorithm shows convergence. 
Over time, the regret value stabilizes, 
indicating that the algorithm 
reaches a steady state of 
accumulated regret. 

\end{document}